\documentclass[conference]{IEEEtran}

\usepackage[breaklinks,colorlinks]{hyperref}
\usepackage{url}
\usepackage{booktabs}
\usepackage{multirow}
\usepackage{amsmath,amssymb,amsfonts,amsthm,nccmath}
\usepackage{mathtools}
\usepackage{nicefrac}
\usepackage{microtype}
\usepackage[ruled]{algorithm2e}
\usepackage{algpseudocode}
\usepackage{enumitem}
\usepackage{subcaption}
\usepackage{wrapfig}
\usepackage{array}
\usepackage{xcolor}
\usepackage{soul}
\usepackage{pifont}

\definecolor{darkred}{HTML}{DC582A}
\definecolor{darkgreen}{HTML}{008264}
\definecolor{pittblue}{HTML}{003594}
\hypersetup{
    linkcolor=pittblue,
    citecolor=pittblue,
}

\newtheorem{definition}{\textbf{Definition}}
\newtheorem{assum}{\textbf{Assumption}}
\newtheorem{conjecture}{\textbf{Conjecture}}
\newtheorem{theorem}{\textbf{Theorem}}[section]
\newtheorem{lemma}[theorem]{\textbf{Lemma}}

\DeclareMathOperator*{\argmax}{arg\,max}
\DeclareMathOperator*{\argmin}{arg\,min}
\newcommand{\tprfpr}[2]{{#1}\scriptsize{@\textcolor{pittblue}{#2\%}}}
\newcommand{\TFB}[2]{\textbf{#1}\scriptsize{@\textcolor{pittblue}{#2\%}}}
\newcommand{\para}[1]{\vspace{5pt}\noindent\textbf{#1}}

\begin{document}

\title{Apollo: \underline{A} \underline{Po}steriori \underline{L}abe\underline{l}-\underline{O}nly Membership Inference Attack Towards Machine Unlearning}

\author{%
    Liou Tang \\
    University of Pittsburgh \\
    Pittsburgh, PA, USA \\
    \texttt{liou.tang@pitt.edu} \\
    \And
    James Joshi \\
    University of Pittsburgh \\
    Pittsburgh, PA, USA \\
    \texttt{jjoshi@pitt.edu} \\
    \And
    Ashish Kundu \\
    Cisco Research \\
    San Jose, CA, USA \\
    \texttt{akundu@cisco.com} \\
}
\author{
    \IEEEauthorblockN{
        \textbf{Liou Tang}\IEEEauthorrefmark{1},
        \textbf{James Joshi}\IEEEauthorrefmark{1},
        \textbf{Ashish Kundu}\IEEEauthorrefmark{2}
    }
    \IEEEauthorblockA{
        \IEEEauthorrefmark{1}University of Pittsburgh, Pittsburgh, PA, USA,\\
        \IEEEauthorrefmark{2}Cisco Research, San Jose, CA, USA,
    }
    Email: liou.tang@pitt.edu, jjoshi@pitt.edu, akundu@cisco.com
}

\maketitle

\begin{abstract}
Machine Unlearning (MU) aims to update Machine Learning (ML) models following requests to remove training samples and their influences on a trained model efficiently without retraining the original ML model from scratch. While MU itself has been employed to provide privacy protection and regulatory compliance, it can also increase the attack surface of the model.
Existing privacy inference attacks towards MU that aim to infer properties of the unlearned set rely on the weaker threat model that assumes the attacker has access to both the unlearned model and the original model, limiting their feasibility toward real-life scenarios. We propose a novel privacy attack, \underline{A} \underline{Po}steriori \underline{L}abe\underline{l}-\underline{O}nly Membership Inference Attack towards MU, \texttt{Apollo}, that infers whether a data sample has been unlearned, following a strict threat model where an adversary has access to the label-output of the unlearned model only. We demonstrate that our proposed attack, while requiring less access to the target model compared to previous attacks, can achieve relatively high precision on the membership status of the unlearned samples.
Our code implementation is available \href{https://github.com/LiouTang/Unlearn-Apollo-Attack}{here}.
\end{abstract}

\section{Introduction}\label{sec:intro}

Machine Unlearning (MU) \cite{Cao2015Towards,Bourtoule2021SISA} has been widely studied as an approach to remove data in response to users' requests from trained Machine Learning (ML) models efficiently without retraining the model from scratch. Various data privacy regulations, e.g., the GDPR \cite{GDPR2016} and CCPA \cite{CCPA2018}, include privacy principles such as \emph{the right to be forgotten}, that allow data subjects to request the removal of their privacy-sensitive or copyrighted data involved in the training of ML models. However, existing literature has also demonstrated that, while MU itself is intended as a method to mitigate privacy (and security) risks in ML models, \emph{the Machine Unlearning process itself} introduces potential privacy and security risks \cite{Liu2025Survey,Tang2024Taxonomy}. The knowledge of an unlearning request, for instance, allows an adversary to observe the discrepancy in the model's behavior before and after the request is executed. The adversary can therefore perform various privacy inference attacks, most notably, Membership Inference Attacks (MIAs) toward MU to infer whether a data sample has been \emph{unlearned} \cite{Chen2021Jepardize,Gao2022Deletion,Lu2022Label,Kurmanji2023SCRUB,Hayes2024Inexact}.

As discussed in Tang and Joshi \cite{Tang2024Taxonomy}, prior works on MIAs towards MU largely rely on the change in behavior between the original model and the unlearned model, and therefore adopt the critical assumption that the adversary has access to both the original as well as the unlearned model.
However, we argue that this assumption is not always feasible, especially in a Machine Learning as a Service (MLaaS) setting where the client has no access to the original model \cite{Hu2023Eraser}. Further, most existing research assumes access to the model posteriors (prediction probabilities) to measure the confidence of the model(s), or focuses on using perturbations to measure the model robustness on a sample \cite{Lu2022Label}. Therefore, existing MIAs towards MU are not generalizable to more real-life scenarios.

In this paper, we propose a novel \underline{A} \underline{Po}steriori \underline{L}abe\underline{l}-\underline{O}nly Membership Inference Attack towards MU, \texttt{Apollo}, that only requires access to the prediction labels of the unlearned model. We show that even with such limited access, an adversary can determine the membership status of the unlearned samples. Our attack exploits the inherent \emph{discrepancy} between an unlearned model and a retrained model, where approximate unlearning induces two types of artifacts in the decision space of the model, namely \textsc{Under-Unlearning} and \textsc{Over-Unlearning}. By observing model behavior on adversarially-constructed inputs for target samples, \texttt{Apollo} identifies the membership status of the target sample by the discrepancy between remaining influence of the data sample in the unlearned model compared to retrained models.
Our key contributions are as follows:

\begin{itemize}
    \item We propose a novel \underline{A} \underline{Po}steriori \underline{L}abe\underline{l}-\underline{O}nly Membership Inference Attack towards Machine Unlearning (\texttt{Apollo}) that considers a strict threat model that assumes only black-box, label-only access to the model after unlearning has been performed.
    \item We identify and formalize two types of discrepancies in approximate unlearning, namely \textsc{Under-Unlearning} and \textsc{Over-Unlearning}, which we utilize to identify unlearning of data samples. We formally establish the existence and boundaries for \textsc{Under-Unlearning} and \textsc{Over-Unlearning}.
    \item We run extensive experiments on different unlearning algorithms/approaches and datasets to show that even with a strict threat model, it is possible to extract membership information of the unlearned samples from an unlearned model alone with high precision. This is a significant threat towards MU that shows direct contradiction with the privacy claims of existing works on unlearning, and emphasizes the need for more careful privacy-preserving unlearning approaches.
\end{itemize}


\section{Literature Review and Preliminaries}\label{sec:lit}

\subsection{Machine Unlearning}\label{ssec:lit-MU}

Machine Unlearning (MU) was first introduced by Cao et al. \cite{Cao2015Towards}, and has evolved into a broad area of study. Loosely defined, MU aims to remove the \emph{influence} of data samples to-be-removed efficiently from a trained model while preserving model performance \cite{Cao2015Towards,Bourtoule2021SISA}.
MU methods target different tiers of model update, including the removal of data influence on (a) model output / behavior, especially on the output in LLMs \cite{Chen2023Forget,Zhang2024Negative}, (b) concepts and knowledge representations \cite{Tan2025Lifting,Maini2024TOFU}, and (c) model parameters \cite{Neel2021Descent,Warnecke2023FT,Guo2020CertRmv}.
MU algorithms can be categorized as (a) \emph{exact unlearning}, where the model is (partially) retrained without the samples to be unlearned \cite{Bourtoule2021SISA,Yan2022Arcane} and (b) \emph{approximate unlearning}, where the model is updated to minimize the influence (or behavioral impact) of the unlearned set \cite{Neel2021Descent,Thudi2022Auditable,Guo2020CertRmv}. We discuss both categories respectively.

\para{Exact Unlearning.} A naive MU algorithm is retaining the model from scratch, ensuring the removal of sample influence in the training process. However, this approach is time consuming and cannot scale. Bourtoule et al. \cite{Bourtoule2021SISA} proposed the first exact unlearning algorithm in which the training set is partitioned into disjoint subgroups, each used to train a sub-model, which are then ensembled. The model owner would therefore only need to retrain sub-models whose training sets contain samples to be unlearned. Yan et al. \cite{Yan2022Arcane} proposed a class-wise partition of the training set to avoid performance loss.

\para{Approximate Unlearning.} Exact unlearning provides strong guarantees of data removal, however, they face high computational costs incurred by retraining, making exact unlearning infeasible for large models or datasets, as well as limitations in model architecture \cite{Hayes2024Inexact}. Therefore, the research community has overwhelmingly adopted approximate unlearning, as recent surveys suggest \cite{Shaik2025Exploring,BJ2025Digital}. In our paper, we focus on \emph{approximate unlearning as the targeted unlearning algorithm}. We provide a brief discussion of baseline and state-of-the-art approximate unlearning algorithms we perform the attacks on in Sec. \ref{ssec:setup}. 

For approximate unlearning, a \emph{strict} definition of MU based on aligning the unlearned model with a retrained model, i.e., the ``golden standard'' of unlearning \cite{Neel2021Descent,Guo2020CertRmv}. Following Chourasia and Shah \cite{Chourasia2023True}, Triantafillou et al. \cite{Triantafillou2024Progress} and Tang and Joshi \cite{Tang2024Taxonomy}, we provide a brief definition of the Machine (Un-)Learning problem as follows:

\begin{definition}[Machine Learning \cite{Chourasia2023True,Tang2024Taxonomy}]\label{def:ML}
    Let $\mathcal{Z} = \mathcal{X} \times \mathcal{Y}$ be the data population. A learning algorithm $\mathcal{A}$ is defined as a mapping:
    \begin{equation}
        \mathcal{A}: \mathcal{Z}^n \to \Theta,
    \end{equation}
    \noindent which takes a dataset $D \in \mathcal{Z}^n$ with $n$ samples as the input and produces a model $\theta = \mathcal{A}(D) \in \Theta$, which is a mapping:
    \begin{equation}
        \theta : \mathcal{X} \to \mathcal{Y}.
    \end{equation}
\end{definition}

\begin{definition}[Machine Unlearning \cite{Chourasia2023True,Tang2024Taxonomy}]\label{def:MU}
    Given a model $\theta = \mathcal{A}(D)$ (Def. \ref{def:ML}), let $D_u \subseteq D$ be the \emph{unlearned set} with $n_u$ samples  and $D_r = D \setminus D_u$ be the \emph{retained set}.\footnote{Note that the assumption of $D_u \subseteq D$ is not always present, as discussed by Tang and Joshi \cite{Tang2024Taxonomy}, we include this requirement such that $D_u \cap D_t = \varnothing$, where $D_t$ is the test set disjoint from the training set drawn from the same population.}
    An unlearning algorithm $\overline{\mathcal{A}}$ is defined as a mapping:
    \begin{equation}
        \overline{\mathcal{A}}: \mathcal{Z}^n \times \mathcal{Z}^{n_u} \times \Theta \to \Theta,
    \end{equation}
    \noindent such that:
    \begin{equation}
        \overline{\mathcal{A}} [D, D_u, \mathcal{A}(D)] \overset{\varepsilon, \delta}{\approx} \mathcal{A}(D_r).
    \end{equation}
    \noindent where the unlearned model $\theta_u$ is \emph{$(\varepsilon, \delta)$-indistinguishable} \cite{Dwork2006DP} from a model trained on the retained set.
\end{definition}

\subsection{Membership Inference Attack}\label{ssec:lit-MIA}

Membership Inference Attacks (MIAs) against ML aim to infer whether a data sample is involved in the training of a model \cite{Shokri2017MIA,Yeom2018Overfit,Carlini2022MIA,Ye2022Enhanced}. MIAs have been widely adopted as an auditing tool for privacy leakage in ML models \cite{Wang2025PrivTool}, and in MU scenarios, whether a data sample has been successfully unlearned / removed \cite{Tang2024Taxonomy}.

\para{Naive MIAs.} In the first proposed MIA against ML proposed by Shokri et al. \cite{Shokri2017MIA}, the authors assumed the adversary has access to the training algorithm $\mathcal{A}$ and the underlying data distribution $\mathcal{Z}$, with which the adversary trains several surrogate (shadow) models to observe the difference in the models' behavior on a sample when it's how samples trained or not trained on the sample. Yeom et al. \cite{Yeom2018Overfit} further formalizes this attack, which identifies the membership advantage of the model on training set samples, where the model has greater confidence than non-traning set members as a result of (inevitable) overfitting on the training data. The adversary trains a binary classifier to distinguish members and non-members.

\para{Likelihood-based MIAs.} Instead of training population-level attack models, the adversary can construct an instance-level MIA in which they conduct hypothesis tests for individual data samples \cite{Carlini2022MIA,Ye2022Enhanced}. In Carlini et al. \cite{Carlini2022MIA}, the adversary trains several shadow models over $\mathcal{Z}$, the shadow models' output probabilities on the target sample $(x, y) \sim \mathcal{Z}$ at label $y$ are then separated into two worlds / hypothesis, where $(x,y)$ is / is not a member of the shadow model training set. The target model's output on $x$ is compared to the worlds, in order to determine the likelihood of it being trained on the target sample. Ye et al. \cite{Ye2022Enhanced} follows a similar intuition, but uses the model confidence on $(x,y)$.

\para{Label-Only MIAs.} Previous discussions of MIAs largely rely on observing disparities in the model outputs / posteriors between members and non-members, and can be easily defended by publishing less information on the data sample. Therefore, another line of attack follows a stricter threat model, where the adversary only observes the prediction label $f_\theta(x) = \argmax_{y} f_\theta(x)_y$. Choquette-Chou et al. \cite{CC2021LabelOnly} first proposed this attack, in which the adversary gradually applies adversarial noise on a data sample to observe the least amount of perturbation needed to change the predicted label on the target data sample, this noise level is used as a surrogate to the model confidence on the sample, where a larger confidence indicates membership status. More recent attacks by Peng et al. \cite{Peng2024OSLO} and Wu et al. \cite{Wu2024YOQO} improves upon this line of attack by reducing the needed number of queries to the target model, we discuss the two proposed attacks in detail in Sec. \ref{ssec:revisit}.

\subsection{Membership Inference Attacks Toward Machine Unlearning}\label{ssec:lit-UMIA}

Building on Sec. \ref{ssec:lit-MIA}, a similar definition for MIAs can be extended to Machine Unlearning, in which an adversary aims to infer whether a data sample has been \emph{unlearned} \cite{Chen2021Jepardize}. We provide a formal definition of Membership Inference Attacks against MU as a security game following Carlini et al. \cite{Carlini2022MIA} and Hayes et al. \cite{Hayes2024Inexact}:

\begin{definition}[Membership Inference Game on Machine Unlearning \cite{Carlini2022MIA,Hayes2024Inexact}]\label{def:MIA-game}
    The game proceeds between the challenger (model owner) $\mathbf{Chal}$ and the adversary $\mathbf{Adv}$:
    \begin{enumerate}
        \item The challenger $\mathbf{Chal}$ trains a model $\theta = \mathcal{A}(D)$ under Def. \ref{def:ML} and subsequently unlearns $D_u \subseteq D$ and produces $\theta_u = \overline{\mathcal{A}} [D, D_u, \mathcal{A}(D)]$ under Def. \ref{def:MU}.
        \item $\mathbf{Chal}$ flips a fair coin for a bit $b \in \{0,1\}$. If $b=0$, the challenger samples $z_{target} \notin D$ from the data domain $\mathcal{Z}$; If $b=1$, the challenger samples $z_{target} \in D_u$.
        \item The adversary $\mathbf{Adv}$ is given query access to $\theta_u$ and any additional knowledge $\mathcal{K}$. For a black-box access attack, $\mathcal{K} = \{ \mathcal{Z}^m \}$; for a white-box access attack, $\mathcal{K} = \{ \mathcal{Z}^m, \theta_u, \theta, \mathcal{A}, \overline{\mathcal{A}}, D_r \}$.\footnote{We follow the general assumption that the adversary can sample surrogate dataset(s) from the same distribution as the target model \cite{Shokri2017MIA,Carlini2022MIA,Hayes2024Inexact}.}
        \item $\mathbf{Adv}$ creates a decision rule $h: \mathcal{Z} \times \mathcal{K} \to \{0,1\}$ and outputs a bit $\hat{b} \gets h^{f_{\theta_u}}(z_{target};\mathcal{K})$.
        \item The adversary wins if $b = \hat{b}$.
    \end{enumerate}
\end{definition}

\begin{table}[t]
    \centering
    \caption{Threat Models of Existing and Proposed MIAs toward MU}
    \label{tab:attack-overview}
    \resizebox{\linewidth}{!}{
    \begin{tabular}{lwc{20pt}wc{20pt}wc{20pt}wc{20pt}}
        \toprule
        \multirow{2.5}{*}{\textbf{Attack Methods}} & \multicolumn{4}{c}{\textbf{Adversary Access}} \\
        \cmidrule(l){2-5}
        & $D$& $\theta$ & $\theta_u$ & ${f_{\theta}(x)}_y$ \\
        \midrule
        Chen et al. \cite{Chen2021Jepardize} & \ding{55} & \ding{51} & \ding{51} & \ding{51} \\
        Gao et al. \cite{Gao2022Deletion} & \ding{55} & \ding{51} & \ding{51} & \ding{51} \\
        Lu et al. \cite{Lu2022Label} & \ding{55} & \ding{51} & \ding{51} & \ding{55} \\
        \midrule
        Naive \texttt{U-MIA}s \cite{Graves2021Amnesiac,Kurmanji2023SCRUB} & \ding{51} & \ding{55} & \ding{51} & \ding{51} \\
        \texttt{U-LiRA} \cite{Hayes2024Inexact} & \ding{55} & \ding{55} & \ding{51} & \ding{51} \\
        \midrule
        \texttt{Apollo} (Ours) & \textcolor{darkred}{\ding{55}} & \textcolor{darkred}{\ding{55}} & \ding{51} & \textcolor{darkred}{\ding{55}} \\
        \bottomrule
    \end{tabular}
    }
\end{table}

\noindent Naive MIAs towards MU (namely \texttt{U-MIA}s) would therefore adopt the same setting as MIAs toward ML, in which the adversary trains a binary classifier with the posteriors of a subset of $D_u$ and a subset of $D_t$ \cite{Graves2021Amnesiac,Kurmanji2023SCRUB}. 
Stronger attacks that adopt the likelihood attack paradigms, e.g., the \texttt{U-LiRA} attack proposed by Hayes et al. \cite{Hayes2024Inexact} computes the likelihood ratio of a sample in the unlearned set.
Other, more sophisticated attacks, as discussed in Sec. \ref{sec:intro}, demonstrate that the change in the models' confidence level on a sample between the original model and the unlearned model is a strong indicator of unlearning \cite{Chen2021Jepardize,Gao2022Deletion}. Further, Lu et al. \cite{Lu2022Label} propose to use the difference of the original and unlearned models' resilience to adversarial perturbation in a label-only access scenario \cite{Lu2022Label}.
In our paper, we focus on an a posteriori, label-only access scenario, a more realistic and stronger threat model (see Sec. \ref{ssec:threatmodel}).
We include an overview of the threat model settings of existing (and our proposed) MIAs toward MU in Table \ref{tab:attack-overview}.

\section{Attacking Machine Unlearning \emph{A Posteriori}}\label{sec:unlearn-posteriori}

\subsection{Threat Model}\label{ssec:threatmodel}

\begin{figure*}[t]
    \centering
    \includegraphics[width=0.8\linewidth]{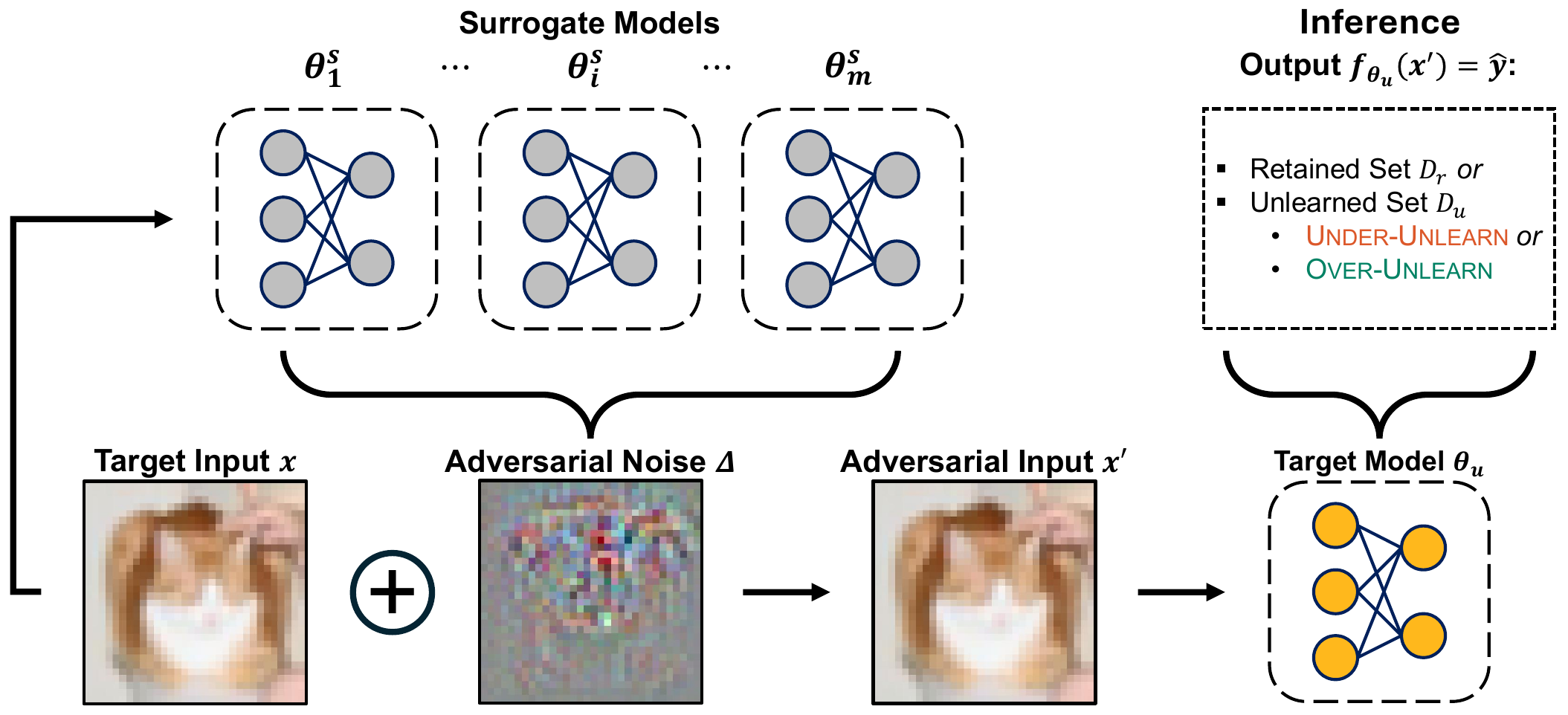}
    \caption{An overview of our proposed \texttt{Apollo} attack. For the target sample whose membership statuses we are interested in, we generate an adversarial input $x^\prime$ under conjectures of \textsc{Under-Unlearning} and \textsc{Over-Unlearning}; the target model's prediction label on the adversarial input is used to infer whether the target sample $x$ is unlearned.}
    \label{fig:overview}
    \vspace{-10pt}
\end{figure*}

\para{Adversary's Ability.} For a MU setting as discussed in Def. \ref{def:MU}, given the unlearned model $\theta_u = \overline{\mathcal{A}} [D, D_u, \mathcal{A}(D)]$, we assume that the adversary has only \emph{black-box} and \emph{label-only} access to the unlearned model $\theta_u$. This means that given an input $\forall z=(x,y)$, the adversary can only know the prediction label output $\hat{y} = \argmax f_{\theta_u}(x)$, without the posterior probabilities. Further, the adversary has no knowledge of $\mathcal{A}$, $\overline{\mathcal{A}}$ or $D$. \emph{We make the explicit assumption that $\overline{\mathcal{A}}$ is an approximate unlearning algorithm}.
This is the strictest (and more challenging) possible threat model for an attack against MU. We follow previous works on MIAs against ML and MU (e.g., \cite{Shokri2017MIA,Peng2024OSLO,Wu2024YOQO,Chen2021Jepardize,Hayes2024Inexact}) in assuming that the adversary has access to a surrogate dataset $D^\prime$ sampled from $\mathcal{Z}$ same as $D$ for the purpose of training shadow models.

\para{Adversary's Goal.} The adversary's goal is to infer whether a data sample $x$ is a member of the unlearned set $D_u$, i.e., the characteristic function $\mathbf{1}_{D_u}(x)$. We aim to demonstrate that, even with minimal access to the target model \cite{Chen2021Jepardize} and more limited knowledge/resources \cite{Hayes2024Inexact}, it is possible to extract privacy-sensitive information from MU. This is a direct rebuttal to the claims of MU as a sound method for data removal.

\para{Intuition.} We design \texttt{Apollo} under the simple intuition that even state-of-the-art Machine Unlearning algorithms produce models that may not align with retrained models, leading to \textsc{Under-Unlearning} and \textsc{Over-Unlearning} observed in the model decision space. We discuss this observation further in Sec. \ref{ssec:revisit}, provide a formal proof for the lower/upper bounds for them in Sec. \ref{ssec:bound} and demonstrate it through a case study in Sec. \ref{ssec:toy-example} and extensive experiments in Sec. \ref{sec:eval}.
This intuition is markedly different from prior MIAs against MU which leverage discrepancy between the original model and unlearned model (\cite{Chen2021Jepardize,Gao2022Deletion,Lu2022Label}).
We show an overview of our attack in Fig. \ref{fig:overview}, which we discuss in detail in Sec. \ref{ssec:design-apollo}.

\subsection{Modifying Label-Only Membership Inference Attacks for Machine Unlearning}\label{ssec:revisit}
We begin by revisiting current state-of-the-art MIAs against ML. 
We can see that both the \texttt{OSLO} attack proposed by Peng et al. \cite{Peng2024OSLO} and the \texttt{YOQO} attack proposed by Wu et al. \cite{Wu2024YOQO} are based on the same assumption: learning leads to \textsc{Over-Learning}; that is, for any $x$ learned, there exists $x^\prime \in \mathcal{X}$ that is classified differently as a result \cite{Wen2023Canary}. While neither Peng et al. \cite{Peng2024OSLO} nor Wu et al. \cite{Wu2024YOQO} have provided a proof for this intuition, it is evidenced by the fact that ML models are able to \emph{generalize} on unseen data. \footnote{A similar notion can be extended to \emph{posterior-based} MIAs, where $x^\prime$ maximizes the divergence between the model outputs on $x^\prime$ when $x \in D$ and $x \notin D$, see Wen et al. \cite{Wen2023Canary}.} We formalize this assumption as follows:

\begin{assum}[Over-Learning \cite{Peng2024OSLO,Wu2024YOQO}]\label{assum:MIA}
    Let $\theta = \mathcal{A}(D)$. Given $(x, y) \in \mathcal{Z}$, there exists $x^\prime \approx x$, such that:
    \begin{equation}
        \left\{
        \begin{aligned}
            &f_{\theta}(x^\prime) = y, &x \in D, \\
            &f_{\theta}(x^\prime) \neq y, &x \notin D.
        \end{aligned}
        \right.
    \end{equation}
    In which learning of $x$ leads to the learning of $x^\prime$, indicated by the change in its prediction label.
\end{assum}

\begin{figure}[t]
    \centering
    \captionsetup[subfigure]{justification=centering}
    \begin{subfigure}[b]{.32\linewidth}
        \centering
        \includegraphics[width=\textwidth]{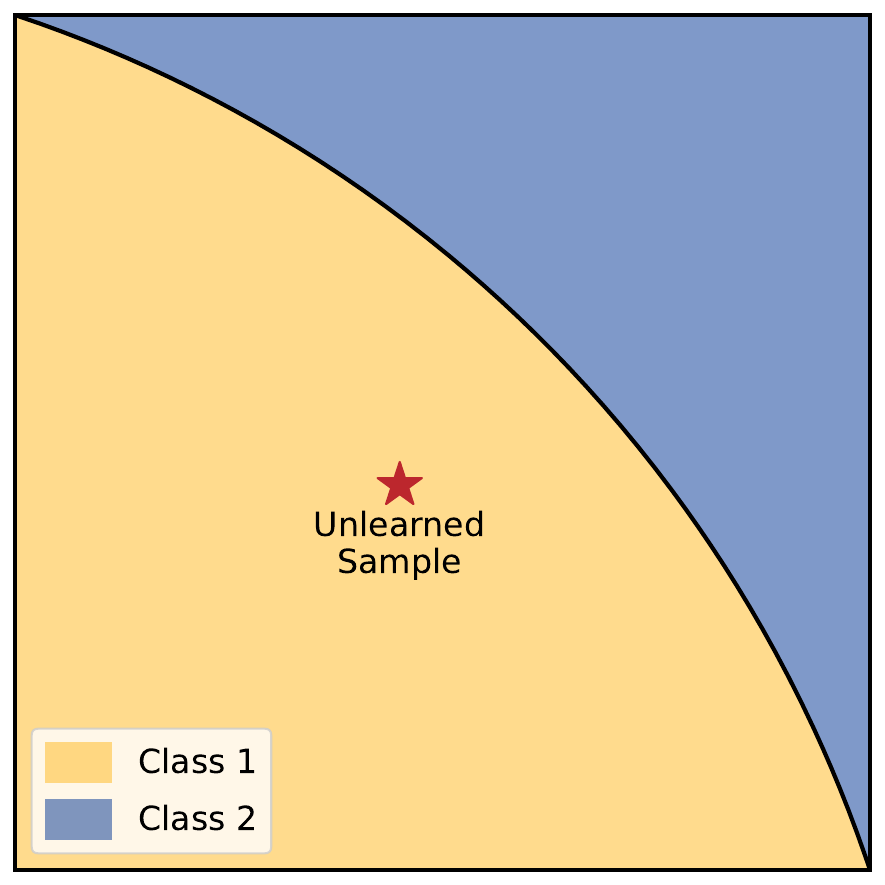}
        \vspace{-10pt}
        \caption{Before\\Unlearning}
        \label{sfig:original}
    \end{subfigure}
    \begin{subfigure}[b]{.32\linewidth}
        \centering
        \includegraphics[width=\textwidth]{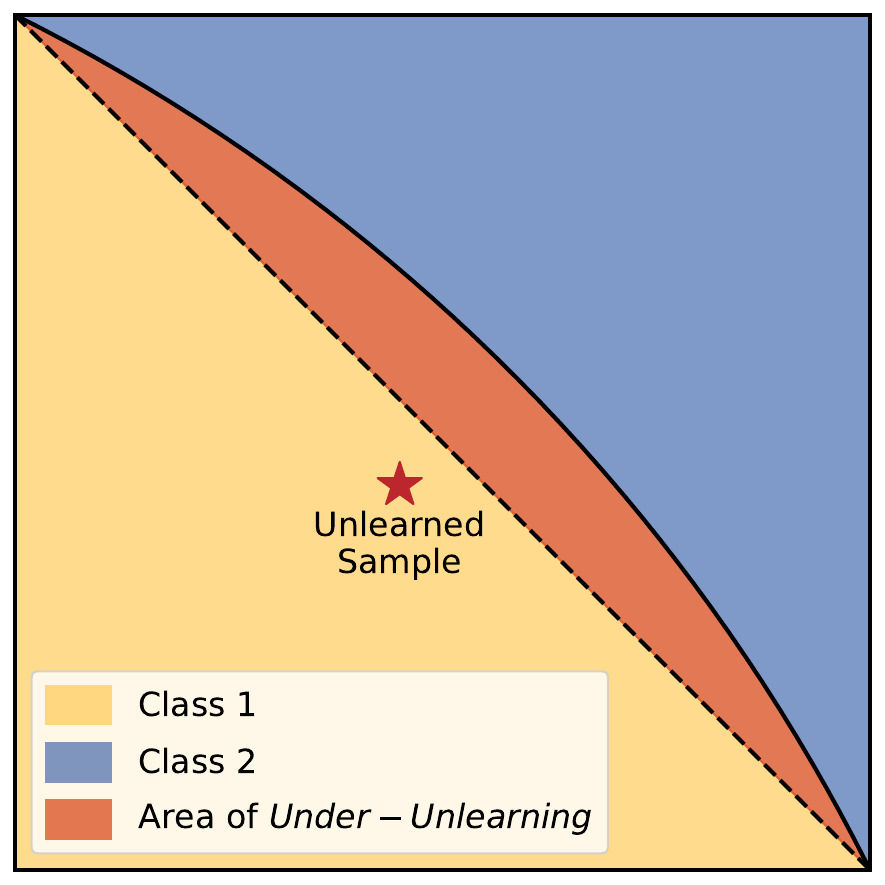}
        \vspace{-10pt}
        \caption{\textsc{Under-\\Unlearning}}
        \label{sfig:under}
    \end{subfigure}
    \begin{subfigure}[b]{.32\linewidth}
        \centering
        \includegraphics[width=\textwidth]{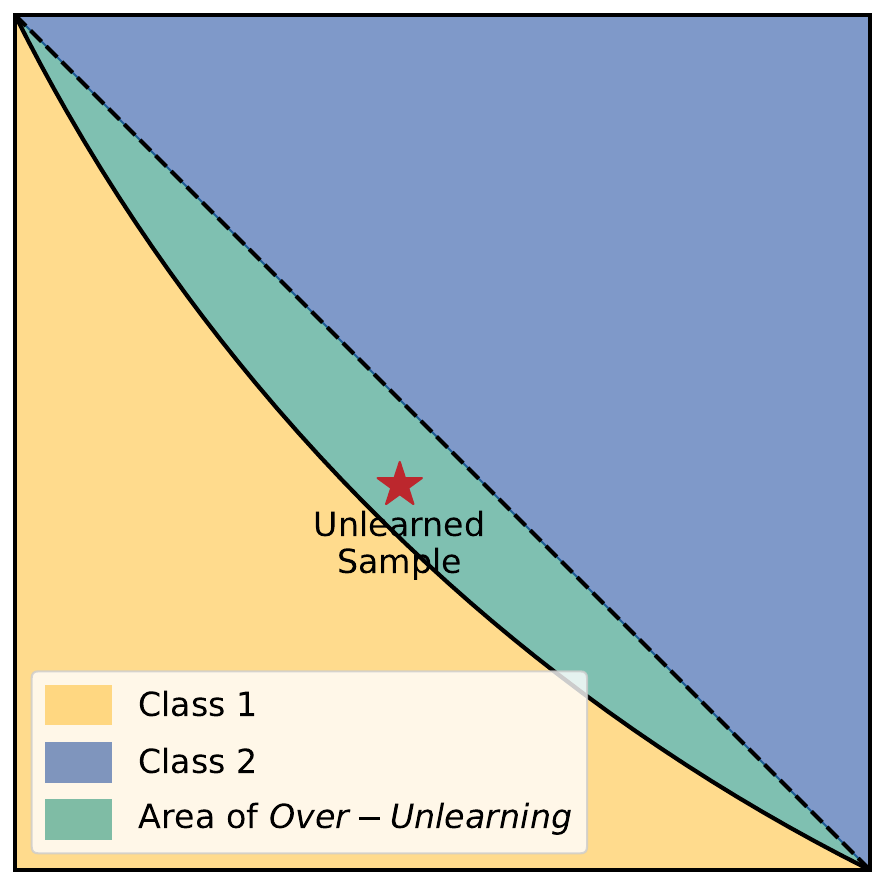}
        \vspace{-10pt}
        \caption{\textsc{Over-\\Unlearning}}
        \label{sfig:over}
    \end{subfigure}
    \caption{We show the dynamics of MU on the decision boundary: (a) Before unlearning (the original model $\theta$), (b) \textcolor{darkred}{\textsc{Under-Unlearning}} (Conj. \ref{conj:under}) and (c) \textcolor{darkgreen}{\textsc{Over-Unlearning}} (Conj. \ref{conj:over}). In each scenario, the solid and dotted line represent the decision boundary of the (approximately) unlearned model $\theta_u$ and a retrained model $\theta_r$, respectively.}
    \label{fig:state}
    \vspace{-10pt}
\end{figure}

We follow a similar yet different intuition in designing \texttt{Apollo}. As shown in Fig. \ref{fig:state}, we present two conjectures: (a) \textsc{Under-Unlearning}, where unlearning retains some information for the unlearned sample due to insufficient removal of influence on the decision landscape of the model and (b) \textsc{Over-Unlearning}, where unlearning imposes performance loss on the retained set due to overzealous change to the decision space. For both conjectures, there exists an \emph{indicator} $x^\prime \in \mathcal{X}$ such that the behavior/prediction labels of the unlearned model and an (exactly) retrained model differ, which we formalize as follows:

\begin{conjecture}[Under-Unlearning]\label{conj:under}
    Following Def. \ref{def:MU}, let $\theta = \mathcal{A}(D)$, $\theta_u = \overline{\mathcal{A}} [D, D_u, \mathcal{A}(D)]$, and $\theta_r = \mathcal{A}(D \setminus D_u)$, where $D_u \subseteq D$. Given an unlearned data sample $(x, y) \in D_u$, there exists an $x^\prime \approx x$, such that:
    \begin{equation}
        \left\{
        \begin{aligned}
            f_{\textcolor{darkred}{\theta}} (x^\prime)   = y &\quad\text{(Learned)} \\
            f_{\textcolor{darkred}{\theta_r}} (x^\prime) \neq y &\quad\text{(Not retained in exact unlearning)} \\
            f_{\textcolor{darkred}{\theta_u}} (x^\prime) = y &\quad\text{(Under-Unlearned in approximate unlearning)}
        \end{aligned}
        \right.
    \end{equation}
    \noindent in which approximate unlearning of $x$ \emph{does not change} the prediction label of $x^\prime$ compared to the original model, while it does compared to the exact unlearning / retraining.
\end{conjecture}

\begin{conjecture}[Over-Unlearning]\label{conj:over}
    Same as Conj. \ref{conj:under}, let $\theta$, $\theta_u$, and $\theta_r$ be the original model, (approximately) unlearned model and retrained model. Given  an unlearned data sample $(x, y) \in D_u$, there exists an $x^\prime\approx x$, such that:
    \begin{equation}
        \left\{
        \begin{aligned}
            f_{\textcolor{darkred}{\theta}} (x^\prime)   = y &\quad\text{(Learned)} \\
            f_{\textcolor{darkred}{\theta_r}} (x^\prime) = y &\quad\text{(Retained in exact unlearning)} \\
            f_{\textcolor{darkred}{\theta_u}} (x^\prime) \neq y &\quad\text{(Over-Unlearned in approximate unlearning)}
        \end{aligned}
        \right.
    \end{equation}
    \noindent in which approximate unlearning of $x$ \emph{changes} the prediction label of $x^\prime$ compared to the original model, while it does not compared to the exact unlearning / retraining.
\end{conjecture}

\noindent For Conj. \ref{conj:under}, $x^\prime$ is depicted in the \textcolor{darkred}{red} region (i.e., area of \textsc{Under-Unlearning}) in Fig. \ref{sfig:under}; for Conj. \ref{conj:over}, $x^\prime$ is depicted in the \textcolor{darkgreen}{green} region (i.e., area of \textsc{Over-Unlearning}) in Fig. \ref{sfig:over}. 
Note that Conj. \ref{conj:under} and Conj. \ref{conj:over} are \emph{not mutually exclusive}, and each speaks to a different aspect of the (inherent) weakness in the unlearning algorithm.

\subsection{Bounding \textsc{Under-} and \textsc{Over-Unlearning}}\label{ssec:bound}

We further ask the following key question about these two conjectures:
\begin{quote}
    Is it possible to derive an lower/upper bound for \textcolor{darkred}{\textsc{Under-Unlearning}} and/or \textcolor{darkgreen}{\textsc{Over-Unlearning}} for a given unlearning process?
\end{quote}
\noindent We formalize this question as follows: 
\begin{quote} 
    Given the original model $\theta = \mathcal{A}(D)$ (Def. \ref{def:ML}), unlearned model $\theta_u = \overline{\mathcal{A}} [D, D_u, \mathcal{A}(D)]$ (Def. \ref{def:MU}.), and a retrained model $\theta_r = \mathcal{A}(D \setminus D_u)$\footnote{Note that $\theta_r$ is not accessible to both the model owner and the adversary. We approximate the behavior of $\theta_r$ with surrogate (shadow) models, see Sec. \ref{ssec:design-apollo}.}, can we calculate: (a) $\|\theta_u - \theta_r\|$, a bound for the change in the model \emph{parameter space}, and (b) $\|x - x^\prime\|$, a bound for the change in the model \emph{decision space}?
\end{quote}
\para{Parameter Space Bound.} As discussed in existing works \cite{Shumailov2021Manipulating,Thudi2022Auditable,Eisenhofer2025Verifiable}, it is possible to construct a scenario where models trained with the same learning algorithm $\mathcal{A}$ on neighboring datasets $D$ and $D^\prime$ that differ in only one element can have the exact same parameters. Golatkar et al. in\cite{Golatkar2021Mixed} propose an approximation of the parameters of $\theta_u$ through first-order-Taylor-series-based decomposition; however, the results rely on the strong assumption that a large-scale underlying dataset is not unlearned; Kurmanji et al. in \cite{Kurmanji2023SCRUB} proposed that a shared information-based definition of MU (a looser definition than the $(\varepsilon, \delta)$-indistinguishability-based definition of MU in Def. \ref{def:MU}) has no upper bound for parameter change. These works on two directions present a possible bound for \textsc{Under-Unlearning} and \textsc{Over-Unlearning} in the model parameter space $\Theta$ to be ${\|\theta - \theta_u\|} \in \left[0, +\infty \right)$.

\para{Decision Space Bound.} While we have demonstrated that we cannot reliably provide a parameter space bound for \textsc{Under-Unlearning} and \textsc{Over-Unlearning}, we can still derive an upper bound for both conjectures in the decision space. We present a short proof as follows.

Let $g_{y,j}(x; \theta) := f_\theta(x)_y - f_\theta(x)_j$, further, define the margin of a sample $(x, y)$ to the decision boundary on $\theta$ as $m_\theta(x) := f_\theta(x)_y - \underset{j \neq y}{\max} f_\theta(x)_j$. We make the following assumptions of standard Lipschitz condition where:

\begin{itemize}
    \item For each $y, j \in \mathcal{Y}$, $j \neq y$, $g_{y,j}(x; \theta)$ is $L_{\textcolor{darkred}{x}}^{(y,j)}$-Lipschitz in $x$;
    \item For each $y, j \in \mathcal{Y}$, $j \neq y$, $g_{y,j}(x; \theta)$ is $L_{\textcolor{darkred}{\theta}}^{(y,j)}$-Lipschitz in $\theta$.
\end{itemize}

\noindent Importantly, this condition may not universally hold for the decision space of ML models \cite{Owhadi2015Brittle,Iglesias2016Bayesian}, though we can assume standard Lipschitz condition \emph{locally} near the data samples and decision boundaries for the model posterior \cite{Hein2017Formal,Tsuzuku2018Lipschitz}.

Denote $L_x := \underset{j \neq y}{\max} L_x^{(y,j)}$, $L_\theta := \underset{j \neq y}{\max} L_\theta^{(y,j)}$. Let $\Delta_u := \| \theta_u - \theta \|$, $\Delta_r := \| \theta_r - \theta \|$ be the changes in model parameters in the approximate / exact unlearning process, we can derive the following lemmas:

\begin{lemma}[Lipschitzness of the Margin]\label{lemma:margin-Lip}
    Let $m_\theta(x)$ be $L_x$-Lipschitz in $x$ and $L_\theta$-Lipschitz in $\theta$; then, for any $x, x^\prime$ and $\theta, \theta^\prime$, we have:
    \begin{equation}\label{eq:lip}
        | m_\theta(x) - m_{\theta^\prime}(x^\prime) | \leq L_x \|x - x^\prime\| + L_\theta \|\theta - \theta^\prime\|.
    \end{equation}
\end{lemma}
\begin{proof}
    Based on the definitions, we have $m_\theta(x) = \underset{j \neq y}{\min} g_{y,j}(x; \theta)$. With the Lipschitz conditions, for each $j \neq y$, we have:
    \begin{equation}
        | m_\theta(x) - m_{\theta}(x^\prime) |
        \leq | g_{y,j}(x; \theta) - g_{y,j}(x^\prime; \theta) |
        \leq L_x \| x - x^\prime \|,
    \end{equation}
    \noindent and,
    \begin{equation}
        | m_\theta(x) - m_{\theta^\prime}(x) |
        \leq | g_{y,j}(x; \theta) - g_{y,j}(x; \theta^\prime) |
        \leq L_\theta \| \theta - \theta^\prime \|.
    \end{equation}
    Based on triangle inequality, we have:
    \begin{equation}\label{eq:lip-proof}
    \begin{aligned}
        | m_\theta(x) - m_{\theta^\prime}(x^\prime) |
        &\leq | m_\theta(x) - m_{\theta}(x^\prime) | + | m_\theta(x) - m_{\theta^\prime}(x) | \\
        & \leq L_x \|x - x^\prime\| + L_\theta \|\theta - \theta^\prime\|.
    \end{aligned}
    \end{equation}
    \noindent Therefore, $m_\theta(x)$ satisfies Lipschitz condition in $x$ and $\theta$.
\end{proof}

\noindent And, from Lemma \ref{lemma:margin-Lip}, we can further derive:

\begin{lemma}[Certified Label-Invariance Radius]\label{lemma:inv}
    If $m_\theta(x) > 0$, $r$ satisfies:
    \begin{equation}
        r < R_{\mathrm{inv}}(\theta \to \theta^\prime, x) :=
        \frac{ m_\theta(x) - L_\theta \| \theta - \theta^\prime \| }{L_x}.
    \end{equation}
    \noindent Then for all $x^\prime$ with $\| x - x^\prime \| \leq r$, $m_{\theta^\prime}(x^\prime) > 0$, i.e., $f_{\theta^\prime}(x^\prime) = y$.
\end{lemma}

Building on Lemma \ref{lemma:margin-Lip} and \ref{lemma:inv}, we propose and prove the following theorems to formally characterize the bounds of \textsc{Under-} and \textsc{Over-Unlearning}, respectively. Here, the bounds are certificates for when (a) the radius below which a change of label is \emph{impossible}, and (b) the radius above which a change of label is \emph{not precluded} by smoothness of the model posterior. For \textsc{Under-Unlearning} (Conj. \ref{conj:under}):

\begin{theorem}[Bounds for \textcolor{darkred}{\textsc{Under-Unlearning}}]\label{th:un-bound}
    For a given $x$, with models $\theta$, $\theta_u$, $\theta_r$, let $r := \| x - x^\prime \|$; if $x^\prime$ satisfies \textsc{Under-Unlearning} (Conj. \ref{conj:under}), we have:
    \begin{equation}
        \left\{
        \begin{aligned}
            r &< \underbrace{\frac{m_\theta(x) - L_\theta \Delta_{\textcolor{darkred}{u}}}{L_x}} 
            =: U_{\mathbf{Un}} \\
            r &\geq \underbrace{\left( \frac{m_\theta(x) - L_\theta \Delta_{\textcolor{darkred}{r}}}{L_x} \right)+}
            =: L_{\mathbf{Un}} \\
        \end{aligned}
        \right.
    \end{equation}
    \noindent The bracelet is non-empty when $(m_\theta(x) - L_\theta \Delta_{r})+ < m_\theta(x) - L_\theta \Delta_{u}$.
\end{theorem}
\begin{proof}
    If $r < U_{\mathbf{Un}}$, Lemma \ref{lemma:inv} certifies that $m_{\theta_u}(x^\prime) > 0$, guaranteeing $f_{\theta_u}(x^\prime) = y$. Conversely, if $r < L_{\mathbf{Un}}$, Lemma \ref{lemma:inv} certifies that $m_{\theta_r}(x^\prime) > 0$, where $\theta_r$ outputs label $y$ at $x^\prime$, contradicting the definition for \textsc{Under-Unlearning} given in Conj. \ref{conj:under}. Therefore, $r \geq L_{\mathbf{Un}}$ is necessary for \textsc{Under-Unlearning} to occur.
\end{proof}

\noindent Similarly, for \textsc{Over-Unlearning} (Conj. \ref{conj:over}), we have:

\begin{theorem}[Bounds for \textcolor{darkgreen}{\textsc{Over-Unlearning}}]\label{th:ov-bound}
    For a given $x$, with models $\theta$, $\theta_u$, $\theta_r$, let $r := \| x - x^\prime \|$; if $x^\prime$ satisfies \textsc{Over-Unlearning} (Conj. \ref{conj:over}), we have:
    \begin{equation}
        \left\{
        \begin{aligned}
            r &< \underbrace{\frac{m_\theta(x) - L_\theta \Delta_{\textcolor{darkred}{r}}}{L_x}} 
            =: U_{\mathbf{Ov}} \\
            r &\geq \underbrace{\left( \frac{m_\theta(x) - L_\theta \Delta_{\textcolor{darkred}{u}}}{L_x} \right)+}
            =: L_{\mathbf{Ov}} \\
        \end{aligned}
        \right.
    \end{equation}
    \noindent The bracelet is non-empty when $(m_\theta(x) - L_\theta \Delta_{u})+ < m_\theta(x) - L_\theta \Delta_{r}$.
\end{theorem}
\begin{proof}
    The proof is same as the proof for Theorem \ref{th:un-bound}.
\end{proof}

Theorem \ref{th:un-bound} and \ref{th:ov-bound} provide theoretical boundaries for \textsc{Under-} and \textsc{Over-Unlearning}, we develop our \texttt{Apollo} attack based on these conjectures in Sec. \ref{ssec:design-apollo}.

\subsection{Designing \texttt{Apollo}}\label{ssec:design-apollo}

\para{Online Attack.} Based on Conj. \ref{conj:under} and \ref{conj:over} and the respective proofs (Theorem \ref{th:un-bound} and \ref{th:ov-bound}), we design \texttt{Apollo} to optimize for $x^\prime$ given the target input $(x, y)$. Denote a family of surrogate (shadow) models $\Theta^s = \left\{\theta^s_i \mid \theta^s_i = \mathcal{A}^s_i(D^s_i) \right\}$, for each $\theta^s_i$, we unlearn $D^{s_u}_i = \{x\}$, and get the corresponding unlearned surrogate (shadow) model $\theta^{s_u}_i$. Further, similar to \texttt{OSLO} \cite{Peng2024OSLO} and \texttt{YOQO} \cite{Wu2024YOQO}, we include the condition that both \textsc{Under-Unlearning} and \textsc{Over-Unlearning} happens \emph{locally}.

In the case of \textsc{Under-Unlearning} (Conj. \ref{conj:under}), the adversarial sample $x^\prime$ for the target sample $(x, y)$ satisfies the conditions of:
\begin{itemize}
    \item[\textbf{(a)}] \textbf{Sensitivity:} $x^\prime$ is classified as class $y$ when it was learned and subsequently unlearned, i.e., \break$x^\prime = \argmin \sum_{x \in D^s_i} \ell(x^\prime; \theta^{s_u}_i)$, in practice, we use Cross Entropy for $\ell$;
    \item[\textbf{(b)}] \textbf{Specificity:} $x^\prime$ is classified \emph{not} as class $y$ when it was not in the model training set, i.e., \break$x^\prime = \argmin \sum_{x \notin D^s_i} \hat{\ell}(x^\prime; \theta^{s}_i)$, in practice, we use $\hat{\ell} = -\ell$;
    \item[\textbf{(c)}] \textbf{Locality:} $ \| x - x^\prime \| < U_{\mathbf{Un}}$ (Theorem \ref{th:un-bound} and \ref{th:ov-bound});
\end{itemize}

\noindent Therefore, we design the loss function for \textsc{Under-Unlearning} (Conj. \ref{conj:under}) as:
\begin{equation}\label{eq:loss-under}
    \ell_{\mathbf{Un}}(x^\prime; x, y, \Theta) =
    \alpha \underbrace{\sum_{x \in D^s_i} \ell(x^\prime; \theta^{s_u}_i)}_{\textbf{(a)}} + 
    \beta  \underbrace{\sum_{x \notin D^s_i} \hat{\ell}(x^\prime; \theta^s_i)}_{\textbf{(b)}}.
\end{equation}
\noindent Here, the locality condition \textbf{(c)} is not directly regulated through the loss function (as we demonstrate in Algorithm \ref{alg:Apollo}). Similarly, for \textsc{Over-Unlearning} (Conj. \ref{conj:over}):
\begin{equation}\label{eq:loss-over}
    \ell_{\mathbf{Ov}}(x^\prime; x, y, \Theta) =
    \alpha \underbrace{\sum_{x \in D^s_i} \textcolor{darkred}{\hat{\ell}}(x^\prime; \theta^{s_u}_i)}_{\textbf{(a)}} + 
    \beta \underbrace{\sum_{x \notin D^s_i} \textcolor{darkred}{\ell}(x^\prime; \theta^s_i)}_{\textbf{(b)}}.
\end{equation}

\para{Offline Attack.} We can further amend our design of \texttt{Apollo} as follows. Note that in Eq. \ref{eq:loss-under} and \ref{eq:loss-over} we need to adjust the unlearned models $\theta^{s_u}_i$ based on every target sample $(x, y)$. This incurs a high computational cost that is infeasible in real-life scenarios. We also design an \emph{offline} version of \texttt{Apollo}, where the adversary can only sample surrogate datasets that are \emph{disjoint} from the target model's training set, i.e., $\forall i, D^s_i \cap D = \varnothing$.
In the offline attack scenario, we can still optimize for the specificity condition \textbf{(b)}; and we substitute the sensitivity condition \textbf{(a)} with the observation that the adversarial samples will achieve better attack accuracy when they are near the \emph{decision boundary} (denoted $\mathrm{DB}$).
The loss functions for the offline \texttt{Apollo} attack(s) would be:
\begin{equation}\label{eq:loss-under-offline}
    \ell^{\mathbf{off}}_{\mathbf{Un}}(x^\prime; x, y, \Theta) =
    \alpha \underbrace{\sum_{i} d(x^\prime, \mathrm{DB})}_{\textbf{(a)}} + 
    \beta  \underbrace{\sum_{i} \textcolor{darkred}{\hat{\ell}}(x^\prime; \theta^s_i)}_{\textbf{(b)}}.
\end{equation}
And:
\begin{equation}\label{eq:loss-over-offline}
    \ell^{\mathbf{off}}_{\mathbf{Ov}}(x^\prime; x, y, \Theta) =
    \alpha \underbrace{\sum_{i} d(x^\prime, \mathrm{DB})}_{\textbf{(a)}} + 
    \beta  \underbrace{\sum_{i} \textcolor{darkred}{\ell}(x^\prime; \theta^s_i)}_{\textbf{(b)}}.
\end{equation}

\para{Adversarial Sample Generation.} We optimize for $x^\prime$ with Stochastic Gradient Descent (SGD), as shown in Algorithm \ref{alg:Apollo}. Here, $\mathcal{B}_{t \cdot \varepsilon}(x) \setminus \mathcal{B}_{(t-1) \cdot \varepsilon}(x) = \{ x^\prime \mid (t-1)\varepsilon < d(x, x^\prime) \leq t\varepsilon \}$ is the space between two balls around $x$ with radius $(t-1) \cdot \varepsilon$ and $t \cdot \varepsilon$. We project adversarial samples outside the ball at each step $t$ to limit the magnitude of each update for the locality condition; the maximum perturbation would be $T \cdot \varepsilon$.
This methodology allows us to adjust the attack precision and recall by selecting different values of $T$, $\varepsilon$, and $\tau$, when lowering $\tau$ and increasing $T \cdot \varepsilon$ allow a high precision.

\begin{algorithm}
    \SetAlgoLined
    \small
    \caption{Adversarial Input Generation in \texttt{Apollo}}\label{alg:Apollo}
    \KwIn{Target model $\theta_u$, target sample $(x, y) \in \mathcal{X} \times \mathcal{Y}$,\break$m$ shadow models $\Theta^s$, $m$ \emph{unlearned} shadow models $\Theta^{s_u}$, search step size $\varepsilon$}
    \KwOut{Adversarial input $x^\prime \in \mathcal{X}$}
    $x^\prime \gets x$\;
    \For{$t \gets 1, \cdots, T$}{
        \For{$i \gets 1, \cdots, m$}{
            $g_{t,i} \gets \nabla_{x^\prime} \textcolor{darkred}{\ell} (x^\prime; x, y, \Theta)$
            \Comment{Substitute for $\ell_{\mathbf{Un}}$, $\ell_{\mathbf{Ov}}$, $\ell^{\mathbf{off}}_{\mathbf{Un}}$, or $\ell^{\mathbf{off}}_{\mathbf{Ov}}$}
        }
        $x^\prime \gets \mathbf{SGD}(x^\prime, \nicefrac{1}{m} \sum_{i=1}^{m} g_{t,i})$\;
        \If{$x^\prime \notin \mathcal{B}_{t \cdot \varepsilon}(x) \setminus \mathcal{B}_{(t-1) \cdot \varepsilon}(x)$}{
            $x^\prime \gets \mathrm{Proj}(x^\prime, \mathcal{B}_{t \varepsilon}(x))$
            \Comment{Limit each update step for locality condition \textbf{(c)}}
        }
        \If{$\nicefrac{1}{m} \sum_{i=1}^m f_{\theta^s_i}{(x^\prime)}_y < \tau$}{
            \Return $x^\prime$
            \Comment{Early stopping}
        }
    }
    \Return $x^\prime$\;
\end{algorithm}

\subsection{A Two Dimensional Example}\label{ssec:toy-example}

\begin{figure}[t]
    \centering
    \captionsetup[subfigure]{justification=centering}
    \begin{subfigure}[b]{.45\linewidth}
        \centering
        \includegraphics[width=\textwidth]{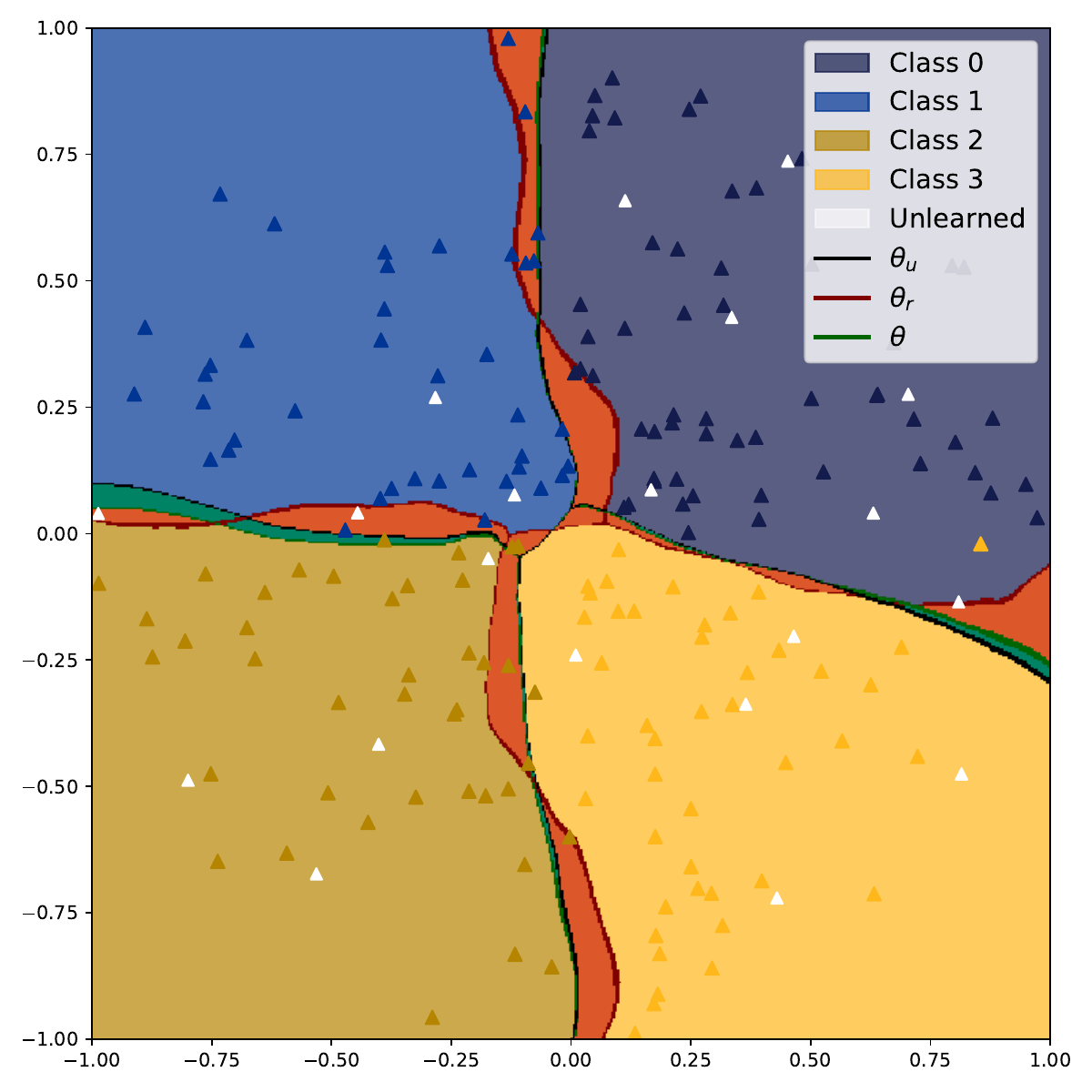}
        \caption{\textsc{Under-Unlearning} and \textsc{Over-Unlearning} on $\mathbb{R}^2$}
        \label{sfig:ToyData}
    \end{subfigure}
    
    \begin{subfigure}[b]{.45\linewidth}
        \centering
        \includegraphics[width=\textwidth]{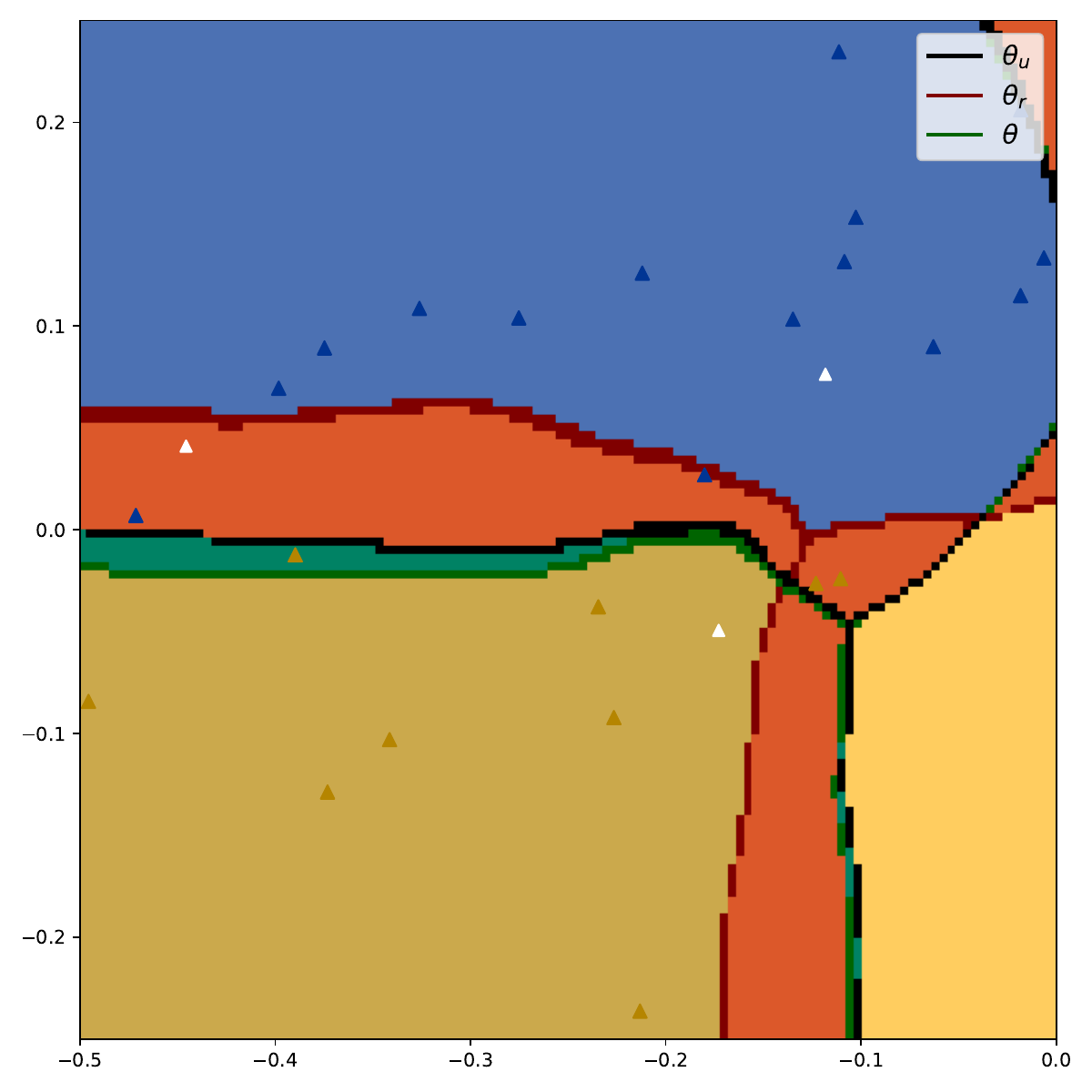}
        \caption{A local region of \textsc{Over-Unlearning} on $\mathbb{R}^2$}
        \label{sfig:ToyData_partial}
    \end{subfigure}
    \begin{subfigure}[b]{.45\linewidth}
        \centering
        \includegraphics[width=\textwidth]{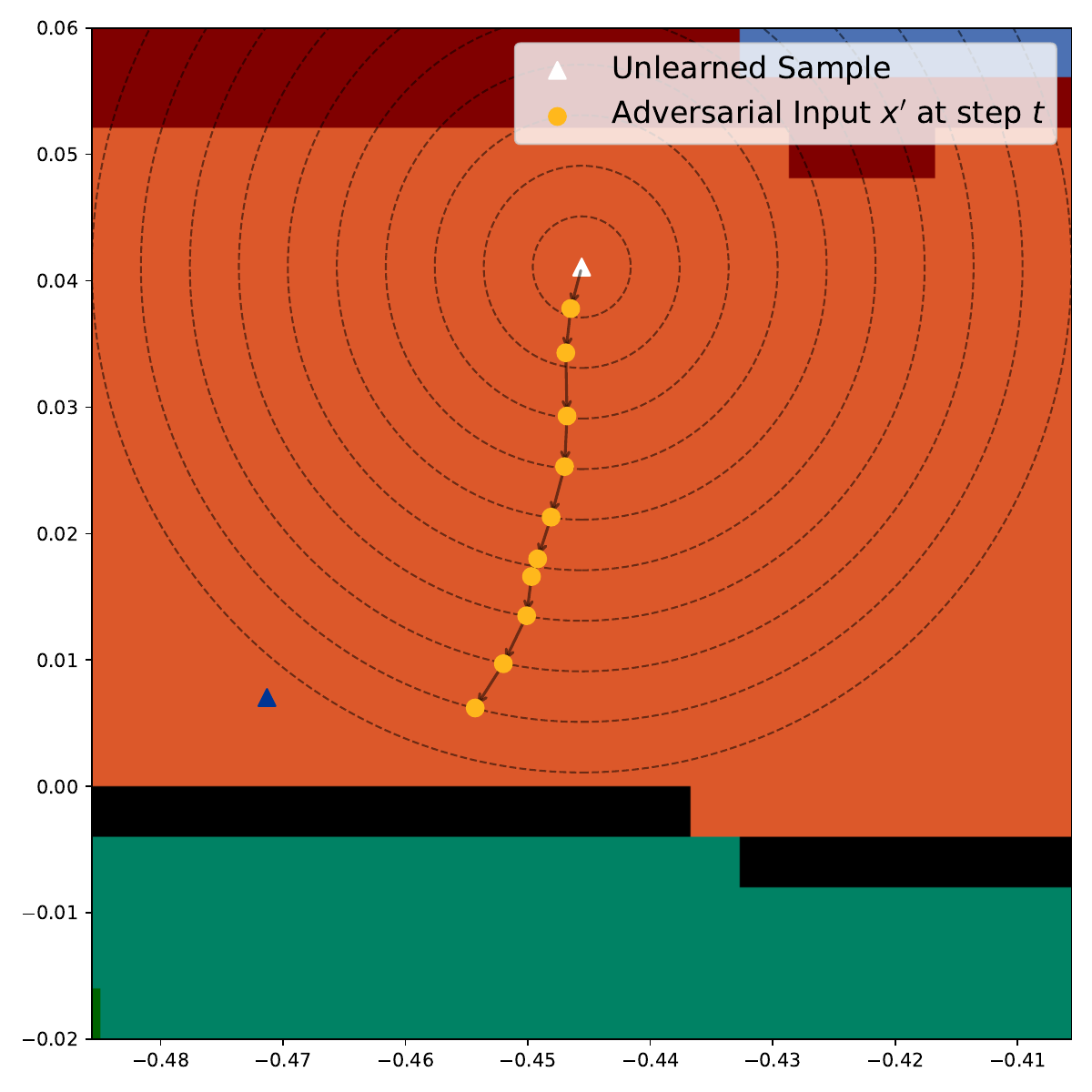}
        \caption{Adversarial Input for an unlearned sample under $\ell_{\mathbf{Un}}$}
        \label{sfig:AdvToyData}
    \end{subfigure}
    \caption{We test Conj. \ref{conj:under} and \ref{conj:over} on a $(x, y) \in \mathbb{R}^2 \times \{0, 1, 2, 3\}$ example. In Fig. \ref{sfig:ToyData}, the four classes and the unlearned set are identified in the legend. Regions of \textcolor{darkred}{\textsc{Under-Unlearning}} and \textcolor{darkgreen}{\textsc{Over-Unlearning}} are colored \textcolor{darkred}{red} and \textcolor{darkgreen}{green}, respectively. In Fig. \ref{sfig:AdvToyData}, the trajectory indicates the adversarial input $x^\prime$ at each iteration.}
    \label{fig:ToyExample}
    \vspace{-10pt}
\end{figure}

Inspired by Haim et al. \cite{Haim2022Reconstruct}, we illustrate our conjectures on a simple example of $(x, y) \in \mathbb{R}^2 \times \{0, 1, 2, 3\}$. We randomly generate $N=500$ samples within the range of $(-1, -1)$ and $(1, 1)$, each class of samples are assigned to a quadrant. We train a target model of 12 layers on 200 samples and $m=16$ shadow models with$|D^s_i| = 200$, the architecture of which we show in Table \ref{tab:toy-net} (Appendix \ref{app:imple}). We randomly unlearn 10\% of the training set with Gradient Ascent (\textbf{GA}) \cite{Thudi2022UnrollSGD}, the results are shown in Fig. \ref{fig:ToyExample}.

As shown in Fig. \ref{sfig:ToyData} and \ref{sfig:ToyData_partial}, MU inevitably leads to both \textcolor{darkred}{\textsc{Under-Unlearning}} (Conj. \ref{conj:under}) and \textcolor{darkgreen}{\textsc{Over-Unlearning}} (Conj. \ref{conj:over}), indicated by the difference in the prediction labels between the unlearned model and a retrained model.

We further demonstrate in this example the effectiveness of our adversarial input generation with Algorithm \ref{alg:Apollo}. We adopt $\ell_{\mathbf{Un}}$ as the loss function, the search process for $x^\prime$ is shown in Fig. \ref{sfig:AdvToyData}. We demonstrate that Algorithm \ref{alg:Apollo} is able to generate $x^\prime$ that satisfies conditions \textbf{(a)} through \textbf{(c)}.

\section{Evaluation}\label{sec:eval}

\subsection{Experimental Setup}\label{ssec:setup}

\para{Hardware.} We train, unlearn, and subsequently perform the attack on all models on a NVIDIA A100 GPU with 40GB memory. The training process require approximately 20 minutes each for the target and shadow models. The time for unlearning process differ with different unlearning algorithms.

\para{Datasets.} Beyond the example on $\mathbb{R}^2$ discussed in Sec. \ref{ssec:toy-example}, we evaluate our attack on commonly employed datasets including CIFAR-10, CIFAR-100 and ImageNet \cite{Krizhevsky2009CIFAR,Deng2009ImageNet}. We train (and subsequently unlearn) the model on a training set of 20,000 images for both CIFAR-10 and CIFAR-100; we train the model on 512,466 images.
We \emph{do not} perform data augmentations on the images to ensure that the unlearned sample is identical to the trained sample \cite{Hayes2024Inexact}.
Both CIFAR-10 and CIFAR-100 are integrated with the \texttt{PyTorch} library \cite{Paszke2019PyTorch}.

\para{Models.} We use ResNet18 \cite{He2016ResNet}, VGG-16 \cite{Simonyan2015VGG} and Swin Transformer \cite{Liu2021SwinT}, for the target and shadow models. For the target model and shadow models, we train them to have an accuracy no lower than 75\% on the test set $D_t$. The target model is trained for 100 epochs with a learning rate of $1\times10^{-4}$ with AdamW optimization algorithm \cite{Loshchilov2019AdamW} and a batch size of 64 on CIFAR-10. The shadow models are trained for 50 epochs for ResNet-18 \cite{He2016ResNet} and VGG-16 \cite{Simonyan2015VGG} from scratch, and 10 epochs for pre-trained Swin Transformer \cite{Liu2021SwinT}. We test our attack on different combinations of the target / surrogate models to determine the generalizability of our attack, which we discuss in Sec. \ref{ssec:ablation}

\para{Unlearning.} We adopt a setting similar to that in Huang et al. \cite{Huang2024Unified}, and test our attack on  6 MU algorithms: Gradient Ascent (\textbf{GA}) \cite{Thudi2022UnrollSGD}, Fine-Tuning (\textbf{FT}) \cite{Warnecke2023FT}, Bad Teacher (\textbf{BT}) \cite{Chundawat2023BadTeach}, \textbf{SCRUB} \cite{Kurmanji2023SCRUB}, \textbf{SalUn} \cite{Fan2024SalUn}, and \textbf{SFR-on} \cite{Huang2024Unified}. We also include Retraining-from-Scratch (\textbf{RT}) as a benchmark: by definition, neither Conj. \ref{conj:under} or \ref{conj:over} should be effective against \textbf{RT}. We randomly unlearn 10\% of the training set. 

The targeted MU algorithms are selected for their representativeness: two baseline unlearning algorithms, \textbf{GA} and \textbf{FT} concerns with only the unlearned set $D_u$ or the retained set $D_r = D \setminus D_u$ respectively; \textbf{BT} uses \emph{knowledge distillation} for unlearning; \textbf{SCRUB} unlearns through maximizing posterior \emph{divergence} between the original model and the unlearned model; \textbf{SalUn} and \textbf{SFR-on} are state-of-the-art unlearning algorithms which share similar intuitions in accounting for the importance (saliency) of different model parameters.

For a more detailed account of the MU algorithms, we refer the readers to their respective papers (\cite{Thudi2022UnrollSGD,Warnecke2023FT,Chundawat2023BadTeach,Kurmanji2023SCRUB,Fan2024SalUn,Huang2024Unified}), as well as the \href{https://github.com/K1nght/Unified-Unlearning-w-Remain-Geometry}{code implementations} by Huang et al. \cite{Huang2024Unified}. 

\para{Metrics.} We simplify the three-way hypothesis test present in MU of $x \in D_u$, $x \in D_r$ and $x \in D_t$ into two categories by sampling 200 samples each from (a) the unlearned set $D_u$ and (b) the test set $D_t$.
We aim to show that our \texttt{Apollo} attack can reliably distinguish between unlearned samples and samples not involved in training.
The main metric we report is the True Positive Rates (TPRs) at low False Positive Rates (FPRs) suggested by Carlini et al. in \cite{Carlini2022MIA}, as a high precision is more valuable to privacy attacks like MIAs.
We also report metrics including precision and recall in our experiments.

\para{Attacks.} We compare our proposed \texttt{Apollo} attack against two previous Membership Inference Attacks toward MU: 
A naive \texttt{U-MIA} employed by Kurmanji et al. \cite{Kurmanji2023SCRUB} and the \texttt{U-LiRA} attack proposed by Hayes et al. \cite{Hayes2024Inexact}. We do not include other attacks listed in Table \ref{tab:attack-overview} (namely \cite{Chen2021Jepardize,Gao2022Deletion,Lu2022Label}), as their assumption of adversarial access to the original model $\theta$ does not align with our threat model.

For \texttt{Apollo}, we adopt a step size $\varepsilon = 1.0$ for $T=50$ rounds of adversarial search for $x^\prime$, in which we utilize $m=32$  shadow models. The loss function (Eq. \ref{eq:loss-under} through \ref{eq:loss-over-offline}) are set to $\alpha=1$ and $\beta=4$. To unlearn each shadow model for the online attack, we unlearn $D^s_i \cap D_{target}$ for each $\theta^{s_u}_i$, in which $D_{target}$ is the 400 samples whose membership status are to be determined.

For \texttt{U-LiRA} \cite{Hayes2024Inexact}, we train and unlearn $m$ shadow models same as \texttt{Apollo}. The \texttt{U-LiRA} attack tests for:
\begin{equation}
    (x,y) \in D_u \iff \frac{\Pr \left( f_{\theta_u}{(x)}_y \mid \mathcal{N}(\mu_{s_u}, \sigma^2_{s_u})  \right)}
    {\Pr \left( f_{\theta_u}{(x)}_y \mid \mathcal{N}(\mu_{s}, \sigma^2_{s})  \right)} > \tau.
\end{equation}
In which $f_{\theta^s_i}{(x)}_y \sim \mathcal{N}(\mu_s, \sigma^2_s)$ and $f_{\theta^{s_u}_i}{(x)}_y \sim \mathcal{N}(\mu_{s_u}, \sigma^2_{s_u})$ are two Gaussian distributions fitted over the prediction probabilities of $x$ on class $y$ with the shadow models $\Theta^s$ and unlearned shadow models $\Theta^{s_u}$. $\tau$ adjusts the precision/recall of the attack. In practice, we use logit-transformed prediction probabilities for $f_{\theta}{(x)}_y$ \cite{Carlini2022MIA,Hayes2024Inexact}.

For the Naive \texttt{U-MIA} described by Kurmanji et al. in \cite{Kurmanji2023SCRUB}, we adopt a Support Vector Machine (SVM) as the binary classifier, which has a regularization parameter of 3 and with radial basis function (RBF) kernel, following the implementation of Fan et al. in \cite{Fan2024SalUn}.


\begin{figure}[t]
    \centering
    \begin{subfigure}{.32\linewidth}
        \includegraphics[height=\textwidth]{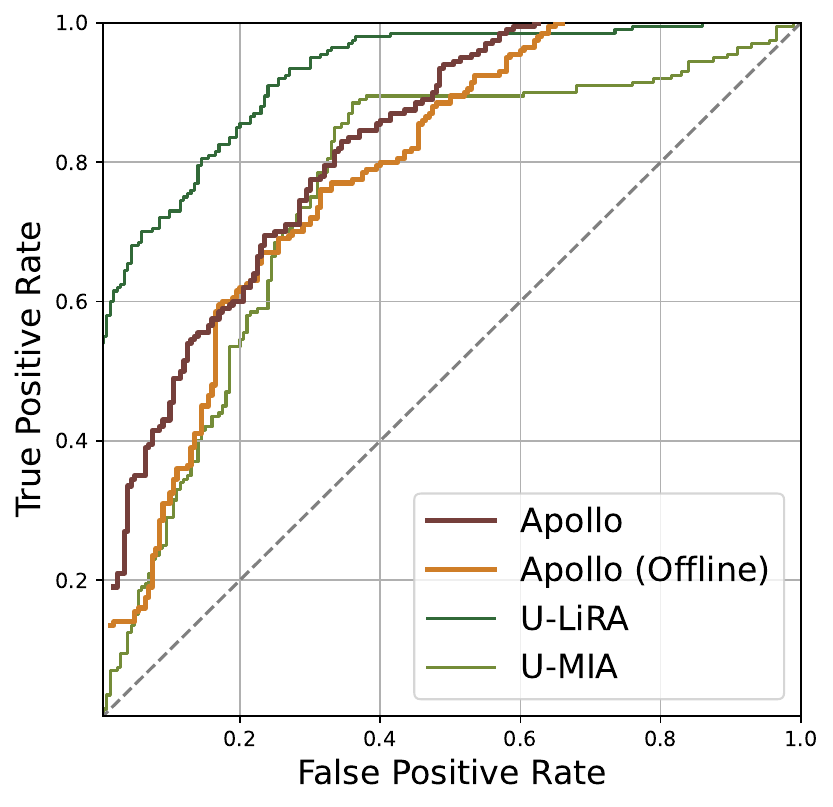}
        \subcaption{\textbf{GA} \cite{Thudi2022UnrollSGD}}
        \label{sfig:roc-GA}
    \end{subfigure}
    \begin{subfigure}{.32\linewidth}
        \includegraphics[height=\textwidth]{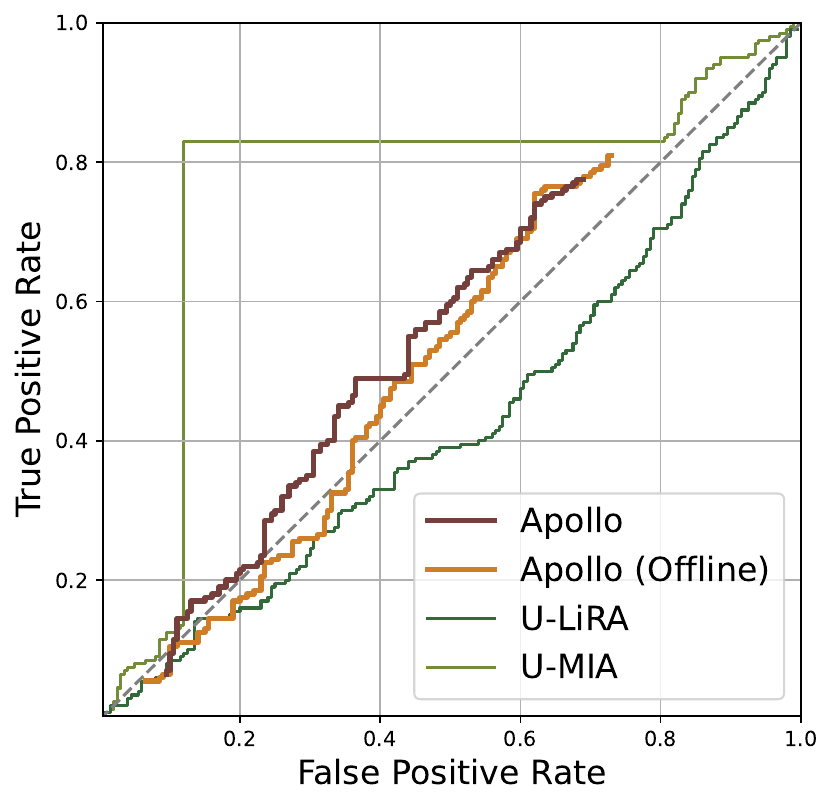}
        \subcaption{\textbf{FT} \cite{Warnecke2023FT}}
        \label{sfig:roc-FT}
    \end{subfigure}
    \begin{subfigure}{.32\linewidth}
        \includegraphics[height=\textwidth]{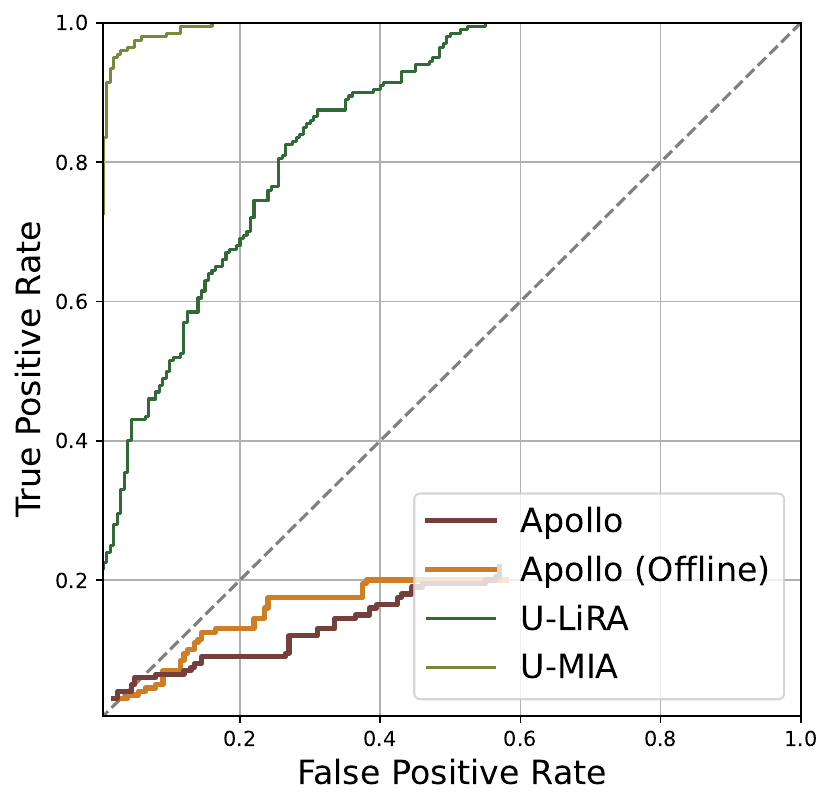}
        \subcaption{\textbf{BT} \cite{Chundawat2023BadTeach}}
        \label{sfig:roc-BT}
    \end{subfigure}
    \begin{subfigure}{.32\linewidth}
        \includegraphics[height=\textwidth]{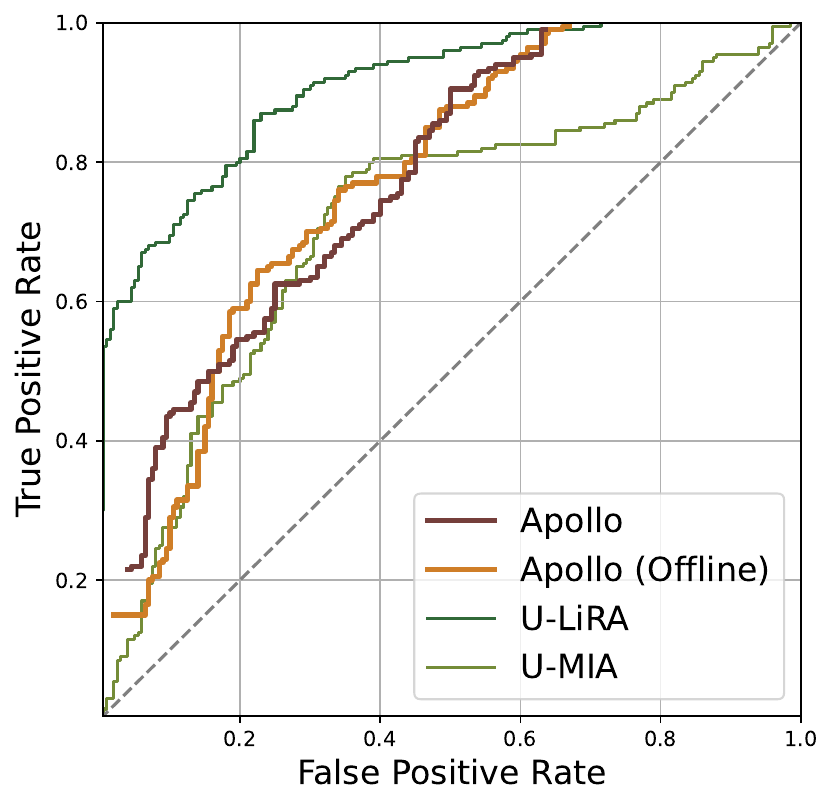}
        \subcaption{\textbf{SCRUB} \cite{Kurmanji2023SCRUB}}
        \label{sfig:roc-SCRUB}
    \end{subfigure}
    \begin{subfigure}{.32\linewidth}
        \includegraphics[height=\textwidth]{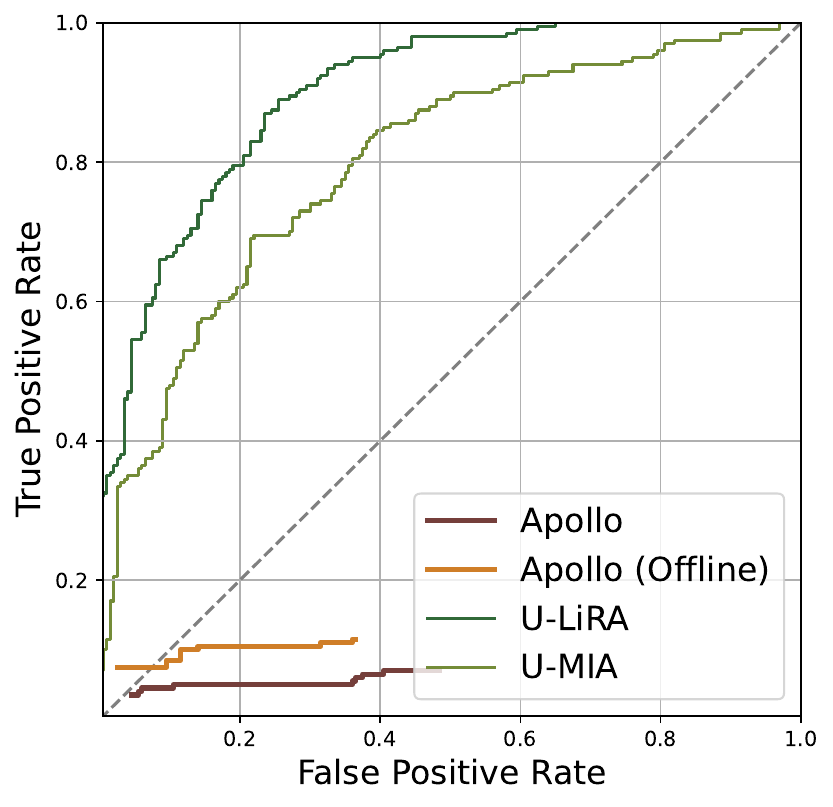}
        \subcaption{\textbf{SalUn} \cite{Fan2024SalUn}}
        \label{sfig:roc-SalUn}
    \end{subfigure}
    \begin{subfigure}{.32\linewidth}
        \includegraphics[height=\textwidth]{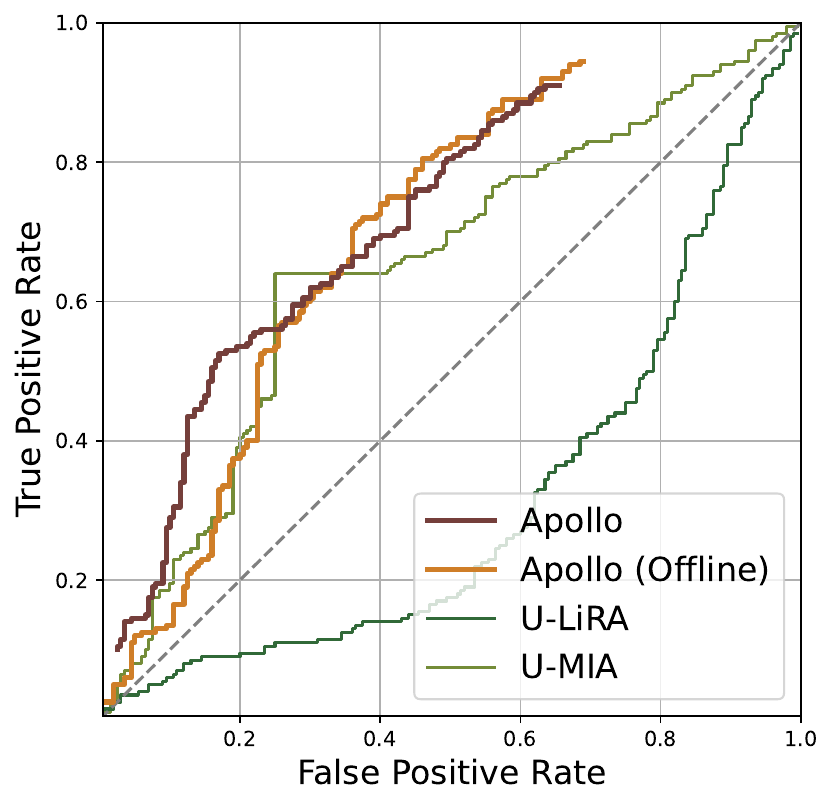}
        \subcaption{\textbf{SFR-on} \cite{Huang2024Unified}}
        \label{sfig:roc-SFRon}
    \end{subfigure}
    \begin{subfigure}{.32\linewidth}
        \includegraphics[height=\textwidth]{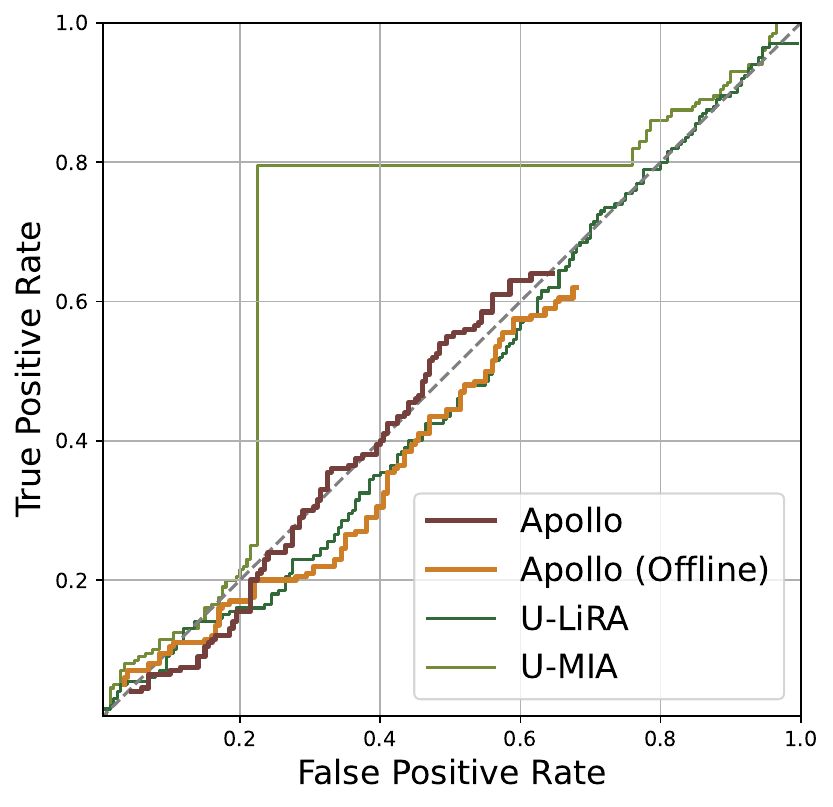}
        \subcaption{\textbf{RT}}
        \label{sfig:roc-RT}
    \end{subfigure}
    \caption{Attack ROCs against various unlearning algorithms.}
    \label{fig:roc}
    \vspace{-10pt}
\end{figure}

\begin{figure}[t]
    \centering
    \begin{subfigure}{.32\linewidth}
        \includegraphics[height=\textwidth]{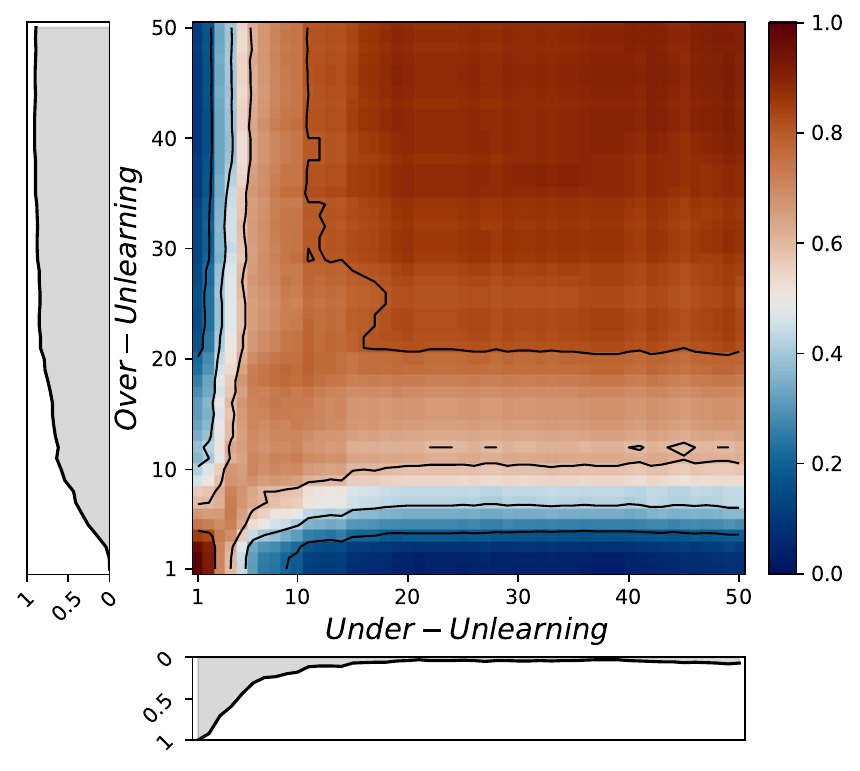}
        \subcaption{\textbf{GA} \cite{Thudi2022UnrollSGD}}
        \label{sfig:dynamic-GA}
    \end{subfigure}
    \begin{subfigure}{.32\linewidth}
        \includegraphics[height=\textwidth]{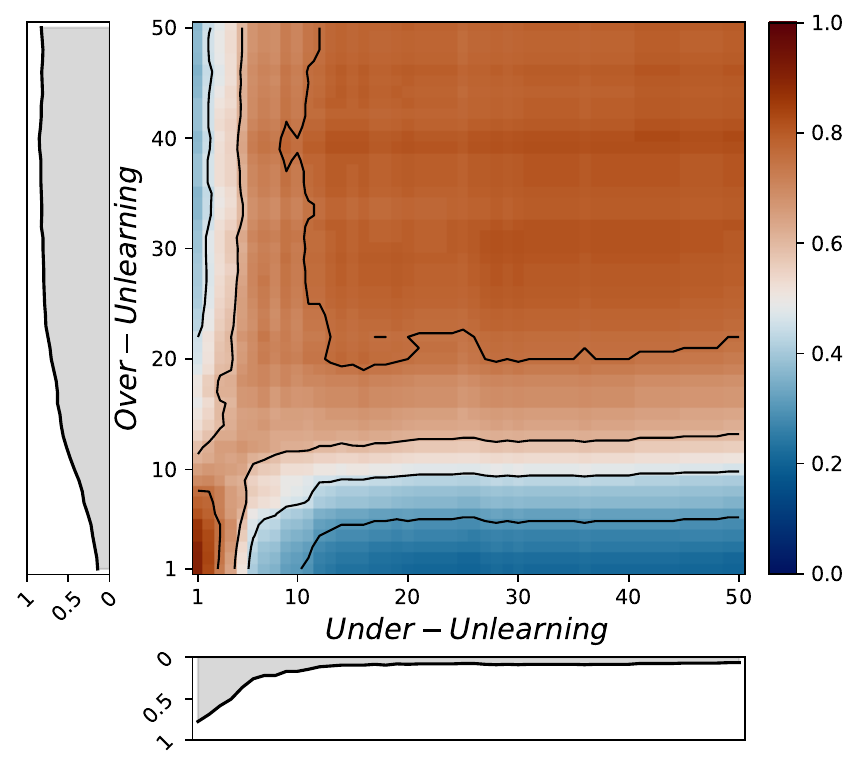}
        \subcaption{\textbf{FT} \cite{Warnecke2023FT}}
        \label{sfig:dynamic-FT}
    \end{subfigure}
    \begin{subfigure}{.32\linewidth}
        \includegraphics[height=\textwidth]{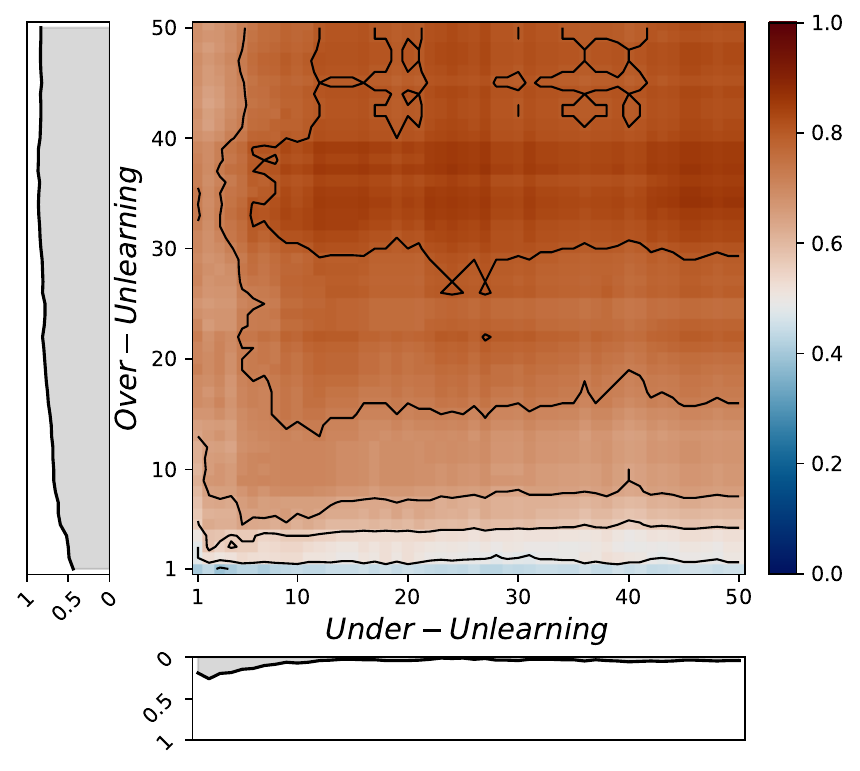}
        \subcaption{\textbf{BT} \cite{Chundawat2023BadTeach}}
        \label{sfig:dynamic-BT}
    \end{subfigure}
    \begin{subfigure}{.32\linewidth}
        \includegraphics[height=\textwidth]{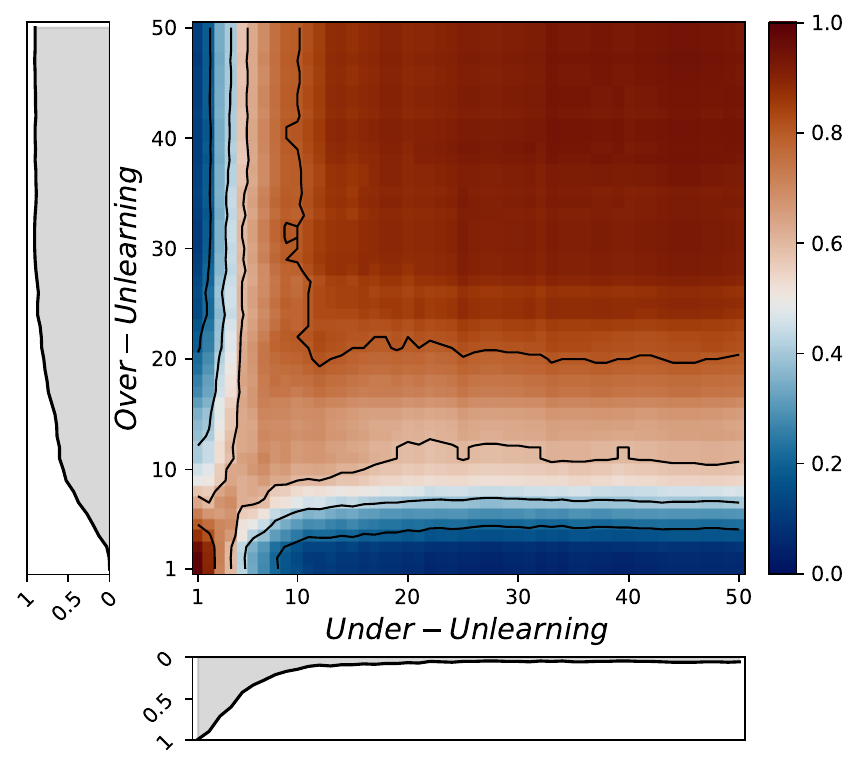}
        \subcaption{\textbf{SCRUB} \cite{Kurmanji2023SCRUB}}
        \label{sfig:dynamic-SCRUB}
    \end{subfigure}
    \begin{subfigure}{.32\linewidth}
        \includegraphics[height=\textwidth]{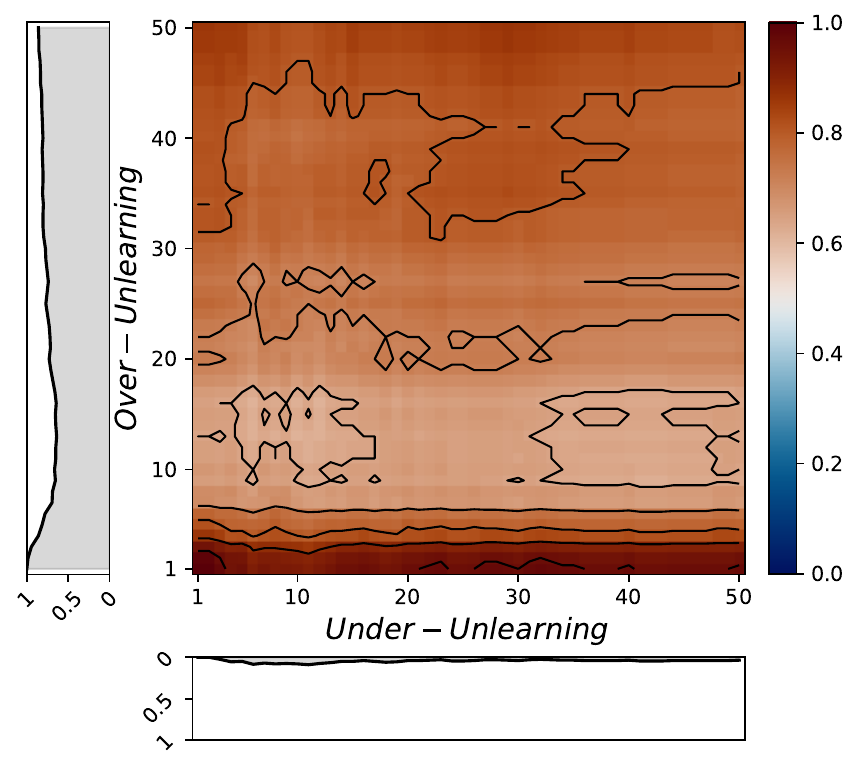}
        \subcaption{\textbf{SalUn} \cite{Fan2024SalUn}}
        \label{sfig:dynamic-SalUn}
    \end{subfigure}
    \begin{subfigure}{.32\linewidth}
        \includegraphics[height=\textwidth]{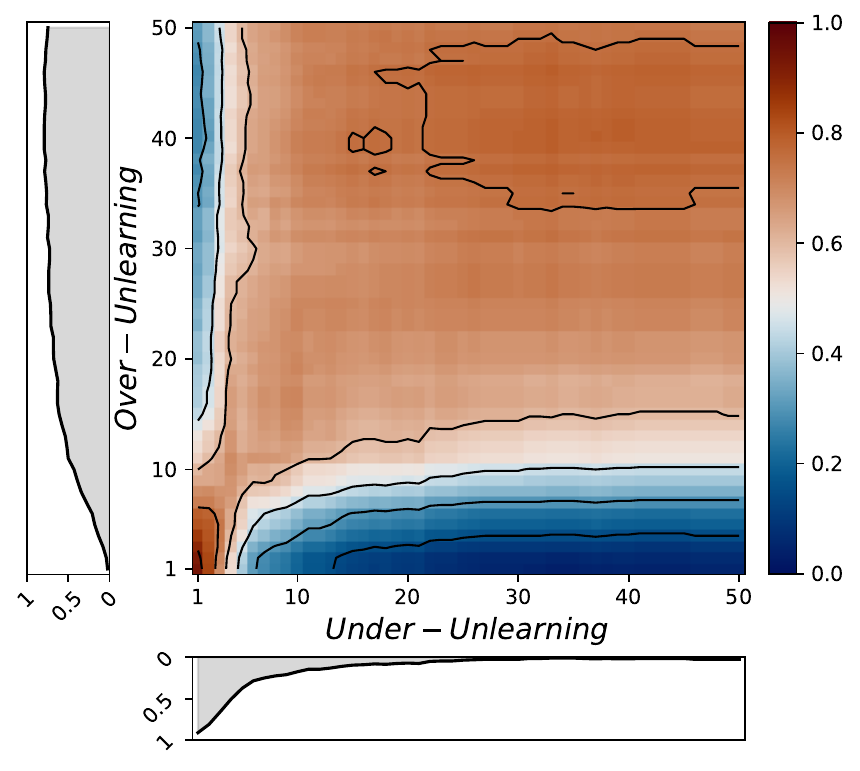}
        \subcaption{\textbf{SFR-on} \cite{Huang2024Unified}}
        \label{sfig:dynamic-SFRon}
    \end{subfigure}
    \begin{subfigure}{.32\linewidth}
        \includegraphics[height=\textwidth]{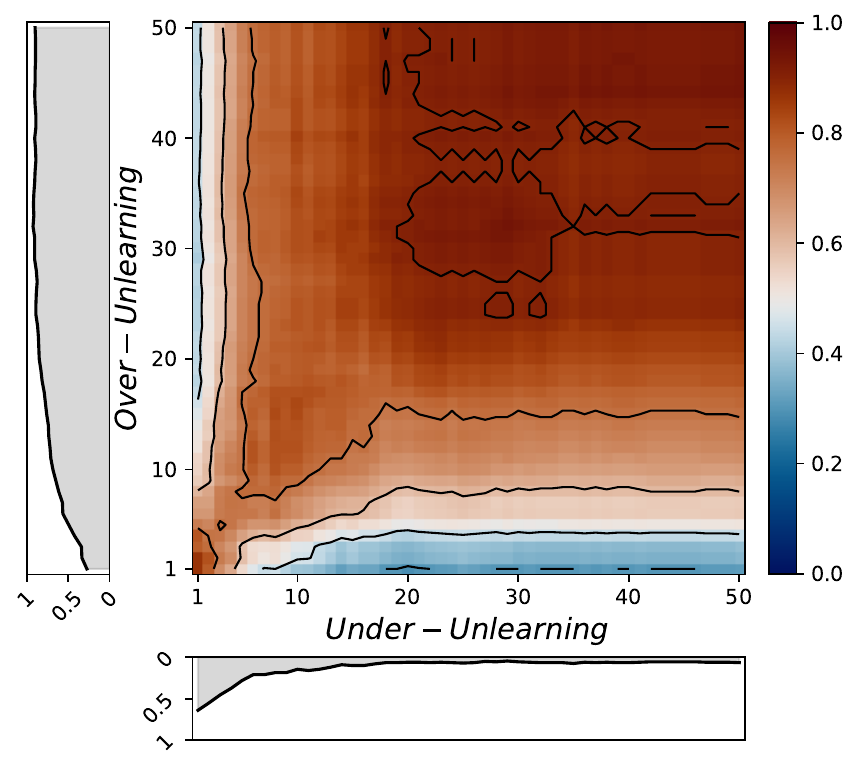}
        \subcaption{\textbf{RT}}
        \label{sfig:dynamic-RT}
    \end{subfigure}
    \caption{Attack TPRs with \textsc{Under-Unlearning} (Conj. \ref{conj:under}) and \textsc{Over-Unlearning} (Conj. \ref{conj:over}) at different step sizes against various unlearning algorithms.}
    \label{fig:dynamic}
    \vspace{-10pt}
\end{figure}

\subsection{Attack Results}\label{ssec:results}

\para{Attack Efficacy.} We show our attack efficacy against different unlearning algorithms compared to previous attacks. We do not adopt the metric of TPRs at a \emph{fixed} FPR across different unlearning algorithms as our search for $x^\prime$ is confined within the ball $\mathcal{B}_{T \cdot \varepsilon}(x)$; therefore, we report TPRs at \emph{lowest possible} FPRs achieved by \texttt{Apollo} and \texttt{Apollo} (offline) for each MU algorithm on CIFAR-10, CIFAR-100 \cite{Krizhevsky2009CIFAR} and ImageNet \cite{Deng2009ImageNet}, shown in Table \ref{tab:attack-low-fpr}, \ref{tab:attack-low-fpr-c100} and \ref{tab:attack-low-fpr-inet}, respectively.

The results demonstrate that our proposed \texttt{Apollo} attack is able to achieve a high TPR at low FPRs, while its performance is limited by the \emph{strict threat model}, it is nevertheless able to achieve a performance comparable to those of \texttt{U-MIA} and \texttt{U-LiRA}, even \emph{superior} than the latter toward \textbf{FT} \cite{Warnecke2023FT} and \textbf{SFR-on} \cite{Huang2024Unified}.
We also note that \texttt{Apollo} is able to achieve stronger performance at lower FPRs on CIFAR-100 (Table \ref{tab:attack-low-fpr-c100}) and ImageNet (Table \ref{tab:attack-low-fpr-inet}) compared to CIFAR-10 (Table \ref{tab:attack-low-fpr}), as each class consists of fewer samples, and a model trained on these datasets is therefore more susceptible to changes in the decision space.

\begin{table*}[t]
    \centering
    \caption{TPRs (\%) at lowest possible \textcolor{pittblue}{FPR}s (\textcolor{pittblue}{\%}), tested on ResNet-18 with CIFAR-10.}
    \renewcommand{\arraystretch}{1.1}
    \label{tab:attack-low-fpr}
    \begin{tabular}{lwc{30pt}wc{30pt}wc{30pt}wc{30pt}wc{40pt}wc{40pt}wc{40pt}}
        \toprule
        \textbf{Attack Method} & \TFB{GA}{6.0} & \TFB{FT}{9.5} & \TFB{BT}{2.5} & \TFB{SCRUB}{4.0} & \TFB{SalUn}{4.5} & \TFB{SFR-on}{2.5} & \TFB{RT}{4.5} \\
        \midrule
        \texttt{U-MIA} \cite{Kurmanji2023SCRUB} & 16.5 & 11.5 & 95.0 & 9.0 & 15.5 & 3.0 & 5.5 \\
        \texttt{U-LiRA} \cite{Hayes2024Inexact} & 68.5 & 6.5 & 28.0 & 6.0 & 20.0 & 2.5 & 4.0 \\
        \midrule
        \texttt{Apollo} & 18.0 & 6.5 & 4.0 & 21.5 & 4.5 & 10.0 & 5.0 \\
        \texttt{Apollo} (Offline) & 16.0 & 6.5 & 3.0 & 15.0 & 7.5 & 5.0 & 7.0 \\
        \bottomrule
    \end{tabular}
\end{table*}

\begin{table*}[ht]
    \centering
    \caption{TPRs (\%) at lowest possible \textcolor{pittblue}{FPR}s (\textcolor{pittblue}{\%}), tested on ResNet-18 with CIFAR-100.}
    \label{tab:attack-low-fpr-c100}
    \begin{tabular}{lwc{30pt}wc{30pt}wc{30pt}wc{30pt}wc{40pt}wc{40pt}wc{40pt}}
        \toprule
        \textbf{Attack Method} & \TFB{GA}{0.5} & \TFB{FT}{1.0} & \TFB{BT}{13.5} & \TFB{SCRUB}{5.0} & \TFB{SalUn}{1.5} & \TFB{SFR-on}{1.5} & \TFB{RT}{1.0} \\
        \midrule
        \texttt{U-MIA} \cite{Kurmanji2023SCRUB} & 7.5 & 0.5 & 48.5 & 17.0 & 8.5 & 2.0 & 1.0 \\
        \texttt{U-LiRA} \cite{Hayes2024Inexact} & 14.5 & 1.0 & 25.0 & 12.5 & 17.0 & 2.0 & 1.5 \\
        \midrule
        \texttt{Apollo} & 15.5 & 2.0 & 50.0 & 41.5 & 5.0 & 0.5 & 1.5 \\
        \texttt{Apollo} (Offline) & 13.0 & 2.0 & 41.5 & 39.0 & 4.5 & 1.0 & 0.5 \\
        \bottomrule
    \end{tabular}
\end{table*}

\begin{table*}[ht]
    \centering
    \caption{TPRs (\%) at lowest possible \textcolor{pittblue}{FPR}s (\textcolor{pittblue}{\%}), tested on ResNet-18 with ImageNet.}
    \label{tab:attack-low-fpr-inet}
    \begin{tabular}{lwc{30pt}wc{30pt}wc{30pt}wc{30pt}wc{40pt}wc{40pt}wc{40pt}}
        \toprule
        \textbf{Attack Method} & \TFB{GA}{1.0} & \TFB{FT}{1.0} & \TFB{BT}{8.5} & \TFB{SCRUB}{4.5} & \TFB{SalUn}{1.0} & \TFB{SFR-on}{1.5} & \TFB{RT}{1.5} \\
        \midrule
        \texttt{U-MIA} \cite{Kurmanji2023SCRUB} & 8.0 & 0.5 & 32.0 & 9.5 & 6.0 & 1.0 & 1.5 \\
        \texttt{U-LiRA} \cite{Hayes2024Inexact} & 12.5 & 0.5 & 24.5 & 8.5 & 8.5 & 2.5 & 2.5 \\
        \midrule
        \texttt{Apollo} & 17.5 & 1.0 & 12.5 & 14.0 & 6.5 & 2.5 & 2.0 \\
        \texttt{Apollo} (Offline) & 13.5 & 0.5 & 8.0 & 14.5 & 2.0 & 2.0 & 1.0 \\
        \bottomrule
    \end{tabular}
\end{table*}

Beyond the attack TPRs at lowest possible FPRs, we also show the ROC curves of different attacks toward different MU algorithms, as shown in Fig. \ref{fig:roc}. Our proposed \texttt{Apollo} attack is able to achieve an overall comparable performance compared to \texttt{U-LiRA} and \texttt{U-MIA} for \textbf{GA}, \textbf{FT}, \textbf{SCRUB} and \textbf{SFR-on}; and in cases able to successfully infer membership status of $x \in D_u$ toward \textbf{SFR-on}, where a likelihood-based \texttt{U-LiRA} attack failed (Fig. \ref{sfig:roc-SFRon}).

\para{Dynamics between \textsc{Under-Unlearning} and \textsc{Over-Unlearning}.}
We next show how \textsc{Under-Unlearning} and \textsc{Over-Unlearning} interacts in different MU algorithms, as well as how the bounds we derived in Theorem \ref{th:un-bound} and \ref{th:ov-bound} perform. We plot the TPRs of \texttt{Apollo} under different search space $\mathcal{B}_{T \cdot \varepsilon} (x)$ for Conj. \ref{conj:under} and \ref{conj:over}. In our experiments, we fix $\varepsilon = 1.0$ and test for steps $t \leq 50$. The sub-figures along the X- and Y-axis represent the TPRs for either conjecture at each step, while the heatmap represents the TPRs where a sample is deemed positive if and only if \emph{one} conjecture holds. The results are shown in Fig. \ref{fig:dynamic}.

Recall our results on TPRs at low FPRs from Table \ref{tab:attack-low-fpr}, here; when our proposed \texttt{Apollo} attack is successful, a clear pattern emerges from the True Positive distributions. For successful attacks like \textbf{GA} (Fig. \ref{sfig:dynamic-GA}) and \textbf{SFR-on} (Fig. \ref{sfig:dynamic-SFRon}), \textsc{Under-Unlearning} and \textsc{Over-Unlearning} are bounded in local regions of $\| x - x^\prime \|$, as evidenced by the low-TPR regions close to the axes, confirming the results in Theorem \ref{th:un-bound} and \ref{th:ov-bound}. For the unsuccessful attack against \textbf{BT} (Fig. \ref{sfig:dynamic-BT}) and \textbf{SalUn} (Fig. \ref{sfig:dynamic-SalUn}), \textsc{Over-Unlearning} is near uniformly distributed, while \textsc{Under-Unlearning} is scarce, we postulate that the regulations of the unlearning algorithms violates the assumption of local Lipschitz condition (Lemma \ref{lemma:margin-Lip}). This indicates that different MU algorithms have distinctively different effects on the decision space of the model, which can be utilized as defense against \texttt{Apollo}.

\subsection{Ablation Studies}\label{ssec:ablation}

\begin{figure}[t]
    \centering
    \begin{subfigure}{.8\linewidth}
        \includegraphics[width=\linewidth]{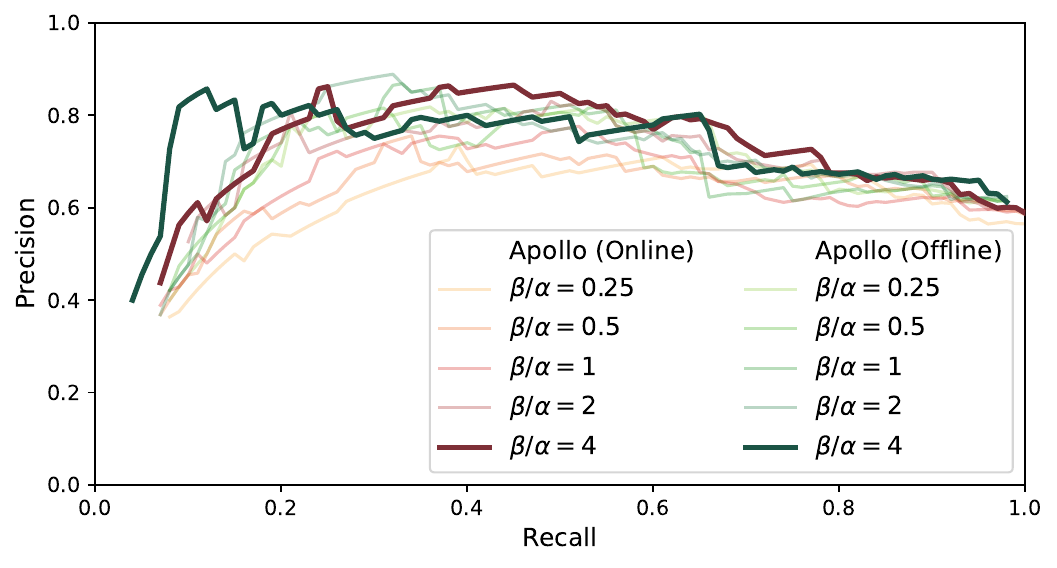}
        \subcaption{Different ratios of $\nicefrac{\beta}{\alpha}$ in the loss function.}
        \label{sfig:prec-rec-ratio}
    \end{subfigure}
    \begin{subfigure}{.8\linewidth}
        \includegraphics[width=\linewidth]{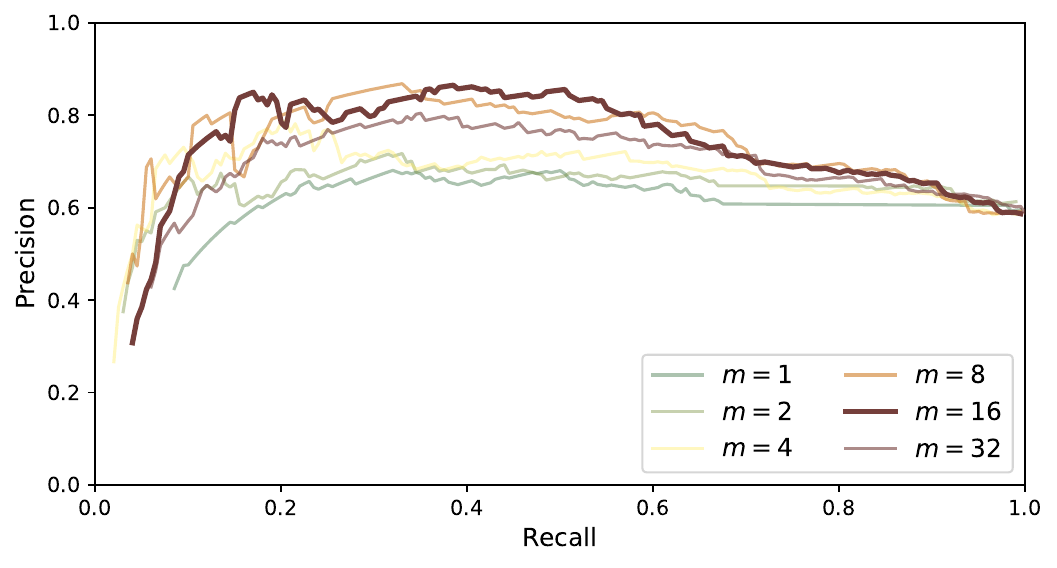}
        \subcaption{Different size of $|\Theta^s| = m$}
        \label{sfig:prec-rec-num}
    \end{subfigure}
    \caption{Precision-Recall trade-off against \textbf{GA} on CIFAR-10, under different hyperparameters used in \texttt{Apollo}.}
    \label{fig:prec-roc}
    \vspace{-10pt}
\end{figure}

\para{Hyperparameters.} As shown in Sec. \ref{ssec:design-apollo}, the loss functions we used for generating the adversarial input $x^\prime$ (Eq. \ref{eq:loss-under} through \ref{eq:loss-over-offline}) follows the general form of $\ell = \alpha \ell_{\text{sensitivity}} + \beta \ell_{\text{specificity}}$. We show the attack precision-recall curves with different ratios of $\nicefrac{\beta}{\alpha}$ in Fig. \ref{sfig:prec-rec-ratio}. We demonstrate that a higher $\nicefrac{\beta}{\alpha}$ ratio, which emphasizes the specificity condition, yields a better precision-recall tradeoff.

\para{Number of Shadow Models.} We study the effectiveness of the number of shadow models, i.e., $|\Theta^s| = m$ on \texttt{Apollo}'s performance, as shown in Fig. \ref{sfig:prec-rec-num}. We observe that while increasing the number of shadow models used can yield a better precision-recall trade-off when $m$ is low ($m \leq 16$), increasing $m$ further results in performance degradation under $m=32$. This is consistent with the observation in Wen et al. \cite{Wen2023Canary} that using too many shadow models will result in $x^\prime$ overfitting to specific shadow models and thus weakening attack utility.

\para{Architecture of Shadow Models.} Another key issue is the training process of the shadow models $\theta^s_i = \mathcal{A}^s_i(D^s_i)$, where we assume that the adversary is able to train shadow models with the same hyperparameters as those of the target model, i.e., $\mathcal{A} = \mathcal{A}^s$. We consider scenarios where the surrogate and target models are trained under different architectures, i.e., $\mathcal{A} \neq \mathcal{A}^s$, as shown in Table \ref{tab:tab:arch-low-fpr}. While employing shadow models with different architectures negatively affects the attack performance, it is still able to preserve a high precision.


\begin{table}[t]
    \centering
    \caption{\texttt{Apollo} performance with different target/shadow model architectures.}
    \label{tab:tab:arch-low-fpr}
    \begin{tabular}{ccccc}
        \toprule
        \textbf{Target Model} & \textbf{Shadow Models} & \textbf{\begin{tabular}{c}TPRs (\%) at lowest \\ possible \textcolor{pittblue}{FPR}s (\textcolor{pittblue}{\%})\end{tabular}} \\
        \midrule
        \multirow{3}{*}{ResNet-18} 
        & ResNet-18 & \tprfpr{18.0}{6.0} \\
        & VGG-16 & \tprfpr{12.0}{6.0} \\
        & Swin-T & \tprfpr{13.5}{6.0} \\
        \midrule
        \multirow{3}{*}{VGG-16} 
        & ResNet-18 & \tprfpr{5.0}{2.5} \\
        & VGG-16 & \tprfpr{5.5}{2.5} \\
        & Swin-T & \tprfpr{6.5}{2.5} \\
        \midrule
        \multirow{3}{*}{Swin-T} 
        & ResNet-18 & \tprfpr{8.5}{4.5} \\
        & VGG-16 & \tprfpr{8.0}{4.5} \\
        & Swin-T & \tprfpr{11.5}{4.5} \\
        \bottomrule
    \end{tabular}
\end{table}

\section{Conclusion}\label{sec:conclusion}

\para{Contributions.} In this paper, we propose \underline{A} \underline{Po}steriori \underline{L}abe\underline{l} \underline{O}nly Membership Inference Attack towards Machine Unlearning, \texttt{Apollo}, that aims to infer the membership status of unlearned data samples from the unlearned model only. We demonstrate empirically that it is possible to extract membership information from a strict threat model against MU, posing a significant privacy threat to existing unlearning methods. The evaluations show that \texttt{Apollo} achieves inference accuracy comparable to those of the state-of-the-art attacks that require more adversarial access than \texttt{Apollo}. MU as a privacy-preserving technique, is still in its infancy; and future research is required to improve both the attacks and defenses for Machine Unlearning techniques.

\para{Limitations.} A limitation of our proposed \texttt{Apollo} attack is that the ability to adjust to different false positive rates is limited by the selection of hyperparameters ($T$, $\varepsilon$, and $\tau$), which is easy for likelihood-based attacks \cite{Kurmanji2023SCRUB,Hayes2024Inexact}. This limitation is inherent in the design of our attack and is the result of adopting a strict threat model. Another limitation of our methodology is the computational cost incurred by training of shadow models, which we address in part in Sec. \ref{ssec:ablation}. Future research is needed for developing more efficient label-only attacks to MU.

\section*{Acknowledgments}

This work is funded by Cisco Research, under proposal id. 95254923.

\bibliographystyle{IEEEtran}
\bibliography{bib}

\clearpage
\onecolumn
\appendices

\section{Additional Implementation Details}\label{app:imple}

\para{Datasets.} We sample our training and surrogate datasets as follows: for CIFAR-10 and CIFAR-100 \cite{Krizhevsky2009CIFAR}, we split the training batch of 50,000 samples into 3 slices, each consisting of (a) 20,000, (b) 20,000 and (c) 10,000 samples, the training set $D$ and test set $D_t$ is then sampled from (a) and (c) respectively. For shadow sets $D^s_i$, in the online version of \texttt{Apollo} and \texttt{U-LiRA} \cite{Hayes2024Inexact}, the shadow sets are sampled from slice (a) and (b); in the offline version of \texttt{Apollo}, the shadow sets are sampled from slice (b) only. For ImageNet \cite{Deng2009ImageNet}, slice (a), (b) and (c) consists of 512,466, 512,466 and 256,235 samples.

\para{Models.} We show the architecture of the small NN used in Section \ref{ssec:toy-example} as follows:

\begin{table}[h]
    \vspace{-5pt}
    \centering
    \caption{Architecture of the small NN used in Section \ref{ssec:toy-example}.}
    \label{tab:toy-net}
    \begin{tabular}{lccc}
        \toprule
        \textbf{Layer} & \textbf{Input Channels} & \textbf{Output Channels} & \textbf{Design} \\
        \midrule
        1 & 2 & 256 & Linear \\
        \midrule
        2-11 & 256 & 256 & \begin{tabular}[c]{@{}c@{}} Linear + ReLU +\\ Batch Norm \end{tabular} \\
        \midrule
        12 & 256 & 4 & Linear \\
        \bottomrule
    \end{tabular}
\end{table}

\vspace{-10pt}
\para{Unlearning.} We list the hyperparameters employed in the targeted unlearning algorithms in Sec. \ref{ssec:setup}:

\begin{table}[h]
    \centering
    \vspace{-5pt}
    \caption{Hyperparameters of the unlearning algorithms employed.}
    \label{tab:param}
    \begin{tabular}{lwc{20pt}wc{40pt}l}
        \toprule
        \textbf{Method} & \textbf{Epochs} & \textbf{Learning Rate} & \textbf{Other Hyperparameters} \\
        \midrule
        \textbf{RT} & 50 & $1\times10^{-4}$ & \begin{tabular}[l]{@{}l@{}}
            constant scheduler,\\ AdamW optimizer
        \end{tabular} \\
        \textbf{GA} & 10 & $2\times10^{-4}$ & cosine scheduler \\
        \textbf{FT} & 10 & $1\times10^{-2}$ & cosine scheduler \\
        \textbf{BT} & 10 & $3\times10^{-3}$ & temperature scalar = 1.0 \\
        \textbf{SCRUB} & 10 & $3\times10^{-4}$ & \begin{tabular}[l]{@{}l@{}}
            constant scheduler,\\ temperature scalar = 4.0,\\ $\alpha = 1\times10^{-3}$
        \end{tabular} \\
        \textbf{SalUn} & 10 & $3\times10^{-4}$ & \begin{tabular}[l]{@{}l@{}}
            constant scheduler,\\ threshold = top-20\%
        \end{tabular} \\
        \textbf{SFR-on} & \textbackslash & \textbackslash & \begin{tabular}[l]{@{}l@{}}
            cosine scheduler,\\
            $T_{out} = 1500$, $\alpha = 1.0$,\\
            $T_{in} = 5$, $\beta^f = 0.25$, $\beta^r = 0.01$,\\
            $\lambda = 0.5$, $\gamma = 1$
        \end{tabular} \\
        \bottomrule
    \end{tabular}
\end{table}







\end{document}